\documentclass{article}
\usepackage{PRIMEarxiv}

\usepackage{float}
\usepackage[table,dvipsnames]{xcolor}
\usepackage{amsmath,amsfonts}
\usepackage[ruled,vlined]{algorithm2e}
\usepackage{graphicx}
\usepackage{mathtools}
\usepackage{mathrsfs}
\usepackage{caption}
\usepackage{amsthm}
\usepackage{amssymb}
\usepackage[colorlinks=true]{hyperref}
\hypersetup{
  linkcolor=black,
  citecolor=black,
  urlcolor=[RGB]{5,115,185}
}
\usepackage{microtype}
\usepackage{multirow}						% cells spanning several rows
\usepackage{tabularx} 						% for larger tables
\usepackage{colortbl}
\usepackage{stmaryrd}
\usepackage{sidecap}
\usepackage{titlesec}
\usepackage{booktabs}
\usepackage[justification=centering]{caption}
\usepackage{array}

\usepackage{PRIMEarxiv}
\usepackage[utf8]{inputenc}
\usepackage[T1]{fontenc}
\usepackage{lmodern}
\usepackage{hyperref}
\usepackage{url}
\usepackage{graphicx}
\usepackage{float}
\usepackage{amsmath,amssymb,amsfonts}
\usepackage{mathrsfs}
\usepackage{bm}    
\usepackage{algpseudocode}
\usepackage{stmaryrd}
\usepackage{nicefrac}
\usepackage{microtype}
\usepackage{mathtools}
\usepackage{authblk}
\usepackage{fancyhdr}
\usepackage{booktabs}
\usepackage{enumitem}
\usepackage{multirow}
\usepackage{footmisc}
\usepackage[skip=6pt]{caption}
\usepackage{subcaption}
\usepackage[table,dvipsnames]{xcolor}
\usepackage{pgfplots}
\pgfplotsset{compat=1.16}
\usepackage{pgfplotstable}
\usepackage{xcolor, colortbl}
\usepackage{algpseudocode}  
\usepackage{titlesec}
\usepackage{amsthm}
\usepackage{tabularx}
\usepackage{booktabs}
\usepackage{tabularx}
\usepackage{array}
\usepackage{pifont}
\newcolumntype{L}{>{\raggedright\arraybackslash}X}   
\newcolumntype{Y}{>{\centering\arraybackslash}X}

\usepackage{xcolor}    
\usepackage{pifont}
\usepackage{graphicx}
\usepackage[justification=centering]{caption}
\usepackage{graphicx}
\usepackage{caption}
\usepackage{tabularx}       % Already required
\usepackage{booktabs}       % For \toprule, \midrule, \bottomrule
\usepackage{float}          % For [H]
\usepackage{listings}

\usepackage[sorting=none, backend=biber]{biblatex} %Imports biblatex package
\definecolor{LightGray}{rgb}{0.8,0.8,0.8}
\definecolor{LightCyan}{rgb}{0.74,0.83,0.9}
\definecolor{DarkBlue}{rgb}{0,0.28,0.67}
\definecolor{blizzardblue}{rgb}{0.67, 0.9, 0.93}
\definecolor{inchworm}{rgb}{0.7, 0.93, 0.36}
\definecolor{coralred}{rgb}{1.0, 0.25, 0.25}
\definecolor{celadon}{rgb}{0.67, 0.88, 0.69}

\DisableLigatures{encoding = *, family = * }

\newcommand{\sherpa}{\href{https://www.sherpa.ai/}{\textcolor{DarkBlue}{Sherpa.ai} }}

\definecolor{DarkColor}{gray}{0.75}			
\definecolor{LightColor}{gray}{0.9}
\definecolor{LightGrey}{rgb}{0.98,0.98,0.98}
\definecolor{DarkGrey}{rgb}{0.83,0.83,0.83}
\definecolor{BaseColor}{rgb}{0.10,0.10,0.20}
\definecolor{TextColor}{RGB}{58,88,119}
\definecolor{LightTextColor}{RGB}{229,233,205}
\definecolor{DarkTextColor}{RGB}{25,29,1}
\definecolor{NeutralBg}{rgb}{0.92,0.92,0.92}
\definecolor{LightYellow}{RGB}{255,255,102}
\definecolor{DarkOrange}{RGB}{255,90,0}
\definecolor{Green}{RGB}{0,128,0}
\definecolor{White}{gray}{1}

\usepackage{etoolbox}
\makeatletter
\patchcmd{\@begintheorem}{\textit}{\textbf}{}{}
\makeatother

\numberwithin{equation}{section}
\numberwithin{subsection}{section}
\usepackage{etoolbox}

\AtBeginEnvironment{tabular}{\small}

\title{Federated Learning for Object Detection: Enabling Collaborative Drone Learning Without Centralizing Data}

\author{%
  {\LARGE \href{https://sherpa.ai/}{Sherpa.ai}}\\
  research@sherpa.ai
}

\fancypagestyle{firstpagestyle}{%
  \fancyhf{}%
  \fancyhead[R]{\includegraphics[scale=1.0]{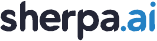}} % logo top-right
  \fancyfoot[C]{\thepage}%
}

\begin{document}

\maketitle
% %%%%%% PUT LOGO IN FIRST PAGE ONLY %%%%%%
\thispagestyle{firstpagestyle}
%%%%%%%%%%%%%%%%%%%%%%%%%%%%%%%%%%%%%%%%%%%

\begin{abstract}

% Object detection is a fundamental capability for AI-driven perception in safety-critical drone and edge-vision systems, including disaster response, operational security environments, infrastructure monitoring and defense applications. Robust model performance in such environments depends on large, continuously updated datasets. However, training high-performing detectors typically requires centralizing aerial imagery, which raises privacy, regulatory, storage, and bandwidth challenges. This is especially problematic in distributed drone deployments, where visual data is generated onboard and is often impractical or undesirable to transfer to a centralized infrastructure.

% In this work, we apply Federated Learning (FL) for object detection, enabling drones to improve a shared model while keeping image data local and private. We implement a federated object detection pipeline using the \sherpa FL platform on the KIIT-MiTA dataset, and compare it with Single-drone and Centralized baselines using mean Average Precision (mAP) at IoU thresholds of 0.50 and 0.50-0.95. In our experiments, the proposed FL approach remains close to Centralized training while dramatically improving over Single-drone training, with the best model (YOLO26 nano) achieving relative gains of $52.89\%$ and $67.80\%$ in mAP@0.50 and mAP@0.50:0.95, respectively. These results show that FL enables scalable, high-performing, and privacy-preserving object detection across distributed drone fleets without data centralization.

Object detection is a fundamental capability for AI-driven perception in safety-critical drone and edge-vision systems, including disaster response, operational security environments, infrastructure monitoring, and defense applications. Robust model performance in such environments depends on large, continuously updated datasets. However, training high-performing detectors typically requires centralizing aerial imagery, which raises privacy, regulatory, storage, and bandwidth challenges. This is especially problematic in distributed drone deployments, where visual data is generated onboard and is often impractical or undesirable to transfer to a centralized infrastructure.

In this work, we apply Federated Learning (FL) for object detection, enabling drones to improve a shared model while keeping image data local and private. We implement a federated object detection pipeline using the \sherpa FL platform on the KIIT-MiTA dataset, and compare it with Single-drone and Centralized baselines using mean Average Precision (mAP) at IoU thresholds of 0.50 and 0.50-0.95. In our experiments, the proposed FL approach remains close to Centralized training while dramatically improving over Single-drone training, with the best lightweight model (YOLO26 nano), suitable for deployment even on very limited edge infrastructure, achieving relative gains of $52.89\%$ and $67.80\%$ in mAP@0.50 and mAP@0.50:0.95, respectively. These results show that FL enables scalable, high-performing, and privacy-preserving object detection across distributed drone fleets without data centralization.

\end{abstract}

\begin{figure}[h]
    \centering
    \includegraphics[width=0.58\linewidth]{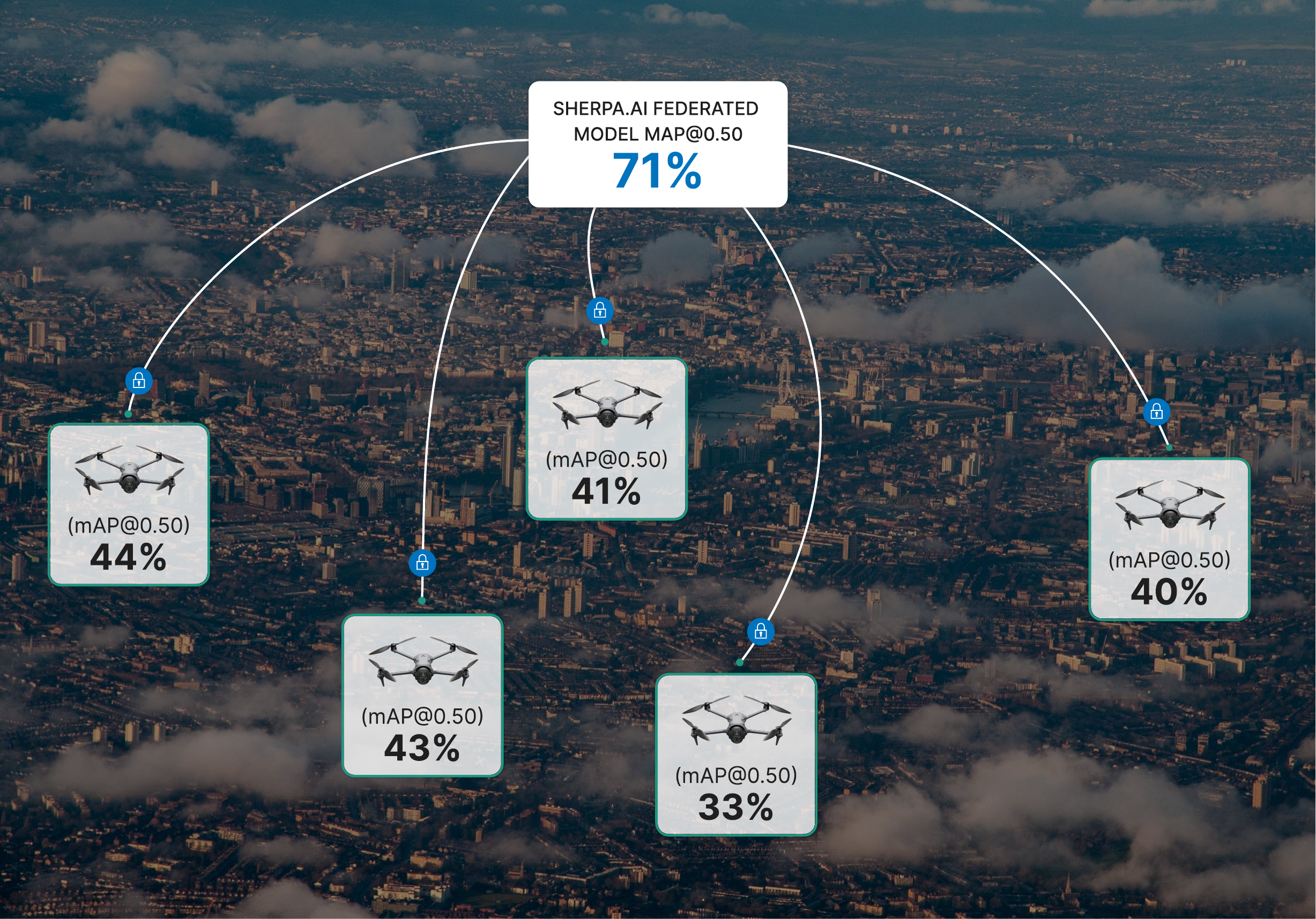}
    \caption{Global map illustrating drone mAP@0.50 for the best federated YOLO model.}
    \label{fig:world_accuracy_comparison}
\end{figure}

% \daniel{The previous abstract should be revised and rewritten once the final results are available.}
% \daniel{I will add here an image to illustrate FL worldwide.}
% \end{abstract}

% \keywords{horizontal federated learning \and object detection \and YOLOv8 \and  Udacity \and Roboflow \and self-driving cars}

%%%%%%%%%%%%%%%%%%%%%%%%%%%%%%%%%%%%%%%%%%%%%%%%%%%%
% \begin{figure}[h]
%     \centering
%     \includegraphics[width=0.54\linewidth]{figs/world_accuracy_comparison.pdf}
%     \caption{Global map illustrating drone mean average precision (mAP) for the best federated YOLO model.}
%     \label{fig:world_accuracy_comparison}
% \end{figure}

\section{Introduction}

Object detection is a fundamental problem in computer vision that involves identifying and localizing instances of semantic objects of a certain class within an image or video. It serves as a crucial component in a wide range of applications such as drone-based surveillance, search and rescue, disaster response, infrastructure inspection, precision agriculture, and robotics. Recent advancements in deep learning have significantly improved the performance of object detection systems. Models such as Faster R-CNN~\cite{ren2015faster}, YOLO~\cite{redmon2016you, redmon2018yolov3, yolov8_ultralytics}, and RetinaNet~\cite{lin2017focal} have become state-of-the-art benchmarks, achieving remarkable accuracy and efficiency in various settings.

In drone vision, object detection is particularly important because unmanned aerial vehicles must operate in dynamic and heterogeneous environments while relying on accurate real-time perception. Compared with ground-based imagery, aerial imagery introduces additional challenges, including small object sizes, viewpoint variations, scale changes due to altitude, cluttered backgrounds, camera motion, and occlusions~\cite{zhu2018vision}. These factors make robust detection from drone platforms especially demanding and increase the importance of training with large, diverse, and well-annotated datasets.

Despite recent progress, several challenges persist in object detection. These include issues related to class imbalance, occlusion, variations in scale and aspect ratio of objects, as well as the need for large, diverse, and high-quality annotated datasets for model training~\cite{zhao2019object}. Furthermore, deploying robust object detection systems in real-world drone scenarios often requires adapting models to heterogeneous data distributions, which can vary significantly across geographic regions, weather conditions, sensors, flight altitudes, and mission profiles. Addressing domain shift and ensuring model generalization under such variability remain key open problems~\cite{wang2023generalized}.

One of the critical obstacles to further improvement is limited data sharing. In many domains, especially those involving critical infrastructure, public safety, industrial inspection, or sensitive geographic information, privacy regulations~\cite{gdpr2016, lopd2018, hipaa1996, CCPA2018} and proprietary concerns prevent centralized aggregation of datasets. In drone-based applications, these constraints are further compounded by limited connectivity, onboard storage restrictions, and the operational cost of transmitting large volumes of image data from edge devices to centralized infrastructure. Consequently, conventional approaches that rely on centralized training with pooled data are often infeasible.

Federated Learning (FL)~\cite{mcmahan2017communication} emerges as a promising paradigm to circumvent these limitations. FL enables multiple participants to collaboratively train a global model without exchanging their local data, thereby preserving data privacy and complying with regulatory constraints (see Figure~\ref{fig:world_accuracy_comparison}). A specific case of FL is Horizontal FL (HFL), where datasets across nodes share the same feature space but differ in the sample space~\cite{yang2019federated}.

Applying HFL to object detection tasks in drone environments offers several tangible benefits. By training models across distributed datasets without requiring raw data exchange, it becomes possible to leverage a wider diversity of aerial scenes, object appearances, and operational conditions~\cite{liu2020fedvision}. This can significantly improve the generalization ability of object detection models, while providing a scalable and privacy-preserving framework that enables continuous model improvement as new data becomes available from different drones or deployment sites without breaching data sovereignty.

\subsection{Motivation}

Drone-based object detection is increasingly relevant in safety-critical scenarios such as military surveillance, disaster response, and infrastructure inspection, where perception systems must operate accurately under real-time and resource-constrained conditions. Robust performance in these environments requires training data that captures diverse aerial viewpoints, object scales, backgrounds, sensors, flight conditions, and operational contexts.

In practice, however, such data is naturally distributed across different drones, operators, or deployment sites. Single-drone training keeps data local but limits each model to the information available at one node, which can reduce generalization. Centralized training can exploit the full dataset but requires transferring raw images to a central server, which may be impractical due to privacy, security, bandwidth, or data governance constraints.

This work is motivated by the following experimental question: Can FL recover most of the benefit of multi-drone training while avoiding raw-image centralization? To answer this, we compare Single-drone, Centralized, and Federated training under a common experimental protocol on the KIIT-MiTA drone object-detection dataset.

\subsection{Contributions}

The main contribution of this work is a controlled experimental evaluation of FL for drone-based object detection. Rather than proposing a new object detector or a new federated optimization algorithm, we focus on assessing whether a standard FL strategy can preserve competitive detection performance in a realistic distributed setting where each node keeps its data local.

Specifically, the contributions of this paper are as follows:
\begin{itemize}
    \item We implement a federated object detection pipeline, enabling collaborative training across four distributed nodes without exchanging raw drone imagery.

    \item We use a non-IID partition of the KIIT-MiTA dataset to evaluate Single-drone, Centralized, and Federated scenarios under a common experimental protocol, enabling a direct comparison of their performance and privacy-utility trade-offs.
    
    \item We benchmark multiple lightweight YOLO nano architectures using both detection and efficiency metrics, and provide aggregate and per-class analyses showing that Federated training consistently improves over Single-drone learning while remaining close to Centralized performance.
\end{itemize}

\section{Problem Formulation}

This section formulates the use case addressed in this work, namely object detection for drones. We first introduce the concept of object detection in the context of aerial perception, and then we formalize the structured prediction problem involved in this task.

\subsection{Object Detection}

\begin{figure}[ht]
	\centering
        \includegraphics[width=1.0\textwidth]{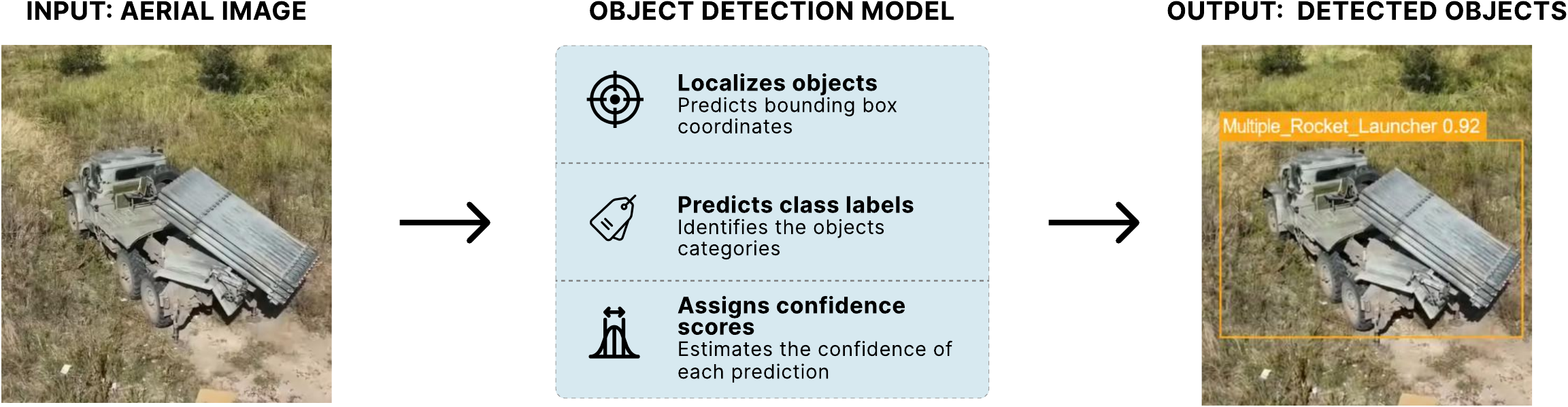}
	\caption{Illustration of the object detection process in aerial drone imagery.}\label{fig:od_processs}
\end{figure}

Object detection is a core perception task in drone-based systems, enabling the platform to identify and localize relevant objects in the observed scene, such as people, vehicles, buildings, and other ground-level or infrastructure elements, depending on the mission context (see Figure~\ref{fig:od_processs}). Unlike standard classification tasks that assign a single label to an entire image, object detection requires predicting both the classes and precise bounding box coordinates for multiple objects within each frame captured by on-board cameras. This structured prediction task is critical for supporting autonomous or assisted drone operations, as it allows the system to perceive dynamic environments, monitor areas of interest, track objects over time, and support decision-making in applications such as surveillance, search and rescue, disaster assessment, and infrastructure inspection.

\subsection{Related Work}
\label{sec:related-work}

This subsection reviews the lines of work most relevant to this paper, namely, centralized and federated approaches to object detection in aerial and drone-based settings. We emphasize prior studies on real-time detectors, aerial-object-detection benchmarks, and privacy-preserving distributed training. This context helps situate the comparative analysis presented in our work on the KIIT-MiTA dataset~\cite{chakrabarty2025kiitmita}.

\subsubsection{Centralized Object Detection for Drones}
Deep-learning-based object detection has progressed rapidly over the last decade. Two-stage detectors such as R-CNN~\cite{girshick2014rich}, Fast R-CNN~\cite{girshick2015fast}, and Faster R-CNN~\cite{ren2015faster} established strong accuracy baselines through region proposals and end-to-end optimization. In parallel, single-stage detectors such as YOLO~\cite{redmon2016you, redmon2018yolov3}, SSD~\cite{liu2016ssd}, and recent Ultralytics YOLO variants~\cite{yolov5_ultralytics,yolov8_ultralytics,yolo11_ultralytics,yolo26_ultralytics} enabled real-time inference by directly regressing bounding boxes and class probabilities. More recently, anchor-free detectors such as FCOS~\cite{tian2019fcos} and transformer-based models such as DETR~\cite{carion2020end} have further broadened the design space for object detection architectures.

When these models are deployed on drones, the task becomes more demanding due to viewpoint changes, camera motion, altitude-dependent scale variation, cluttered backgrounds, and the prevalence of very small objects. Recent surveys on aerial object detection highlight that these factors make standard ground-view benchmarks insufficient for evaluating perception systems intended for UAV platforms~\cite{zhu2018vision}. In addition, real-time aerial detection must often satisfy tight compute and latency constraints on edge hardware, which further motivates lightweight yet accurate detector designs~\cite{leng2024recent}.

A second important line of work concerns dataset design. Traditional benchmarks such as COCO~\cite{lin2014microsoft} and PASCAL VOC~\cite{everingham2010pascal} remain valuable for general-purpose evaluation, but they do not fully reflect the geometric and semantic properties of aerial imagery. Drone-specific datasets have therefore become increasingly important for realistic benchmarking. In particular, Chakrabarty et al.~\cite{chakrabarty2025drones} study real-time military target surveillance from drones using several YOLO-based detection and tracking pipelines. Their results show that modern one-stage detectors are effective for aerial military-scene perception, with YOLO11 small combined with ByteTrack providing the best overall performance. These findings support the use of recent YOLO-based detectors as strong baselines for drone-based military target detection.

\subsubsection{Federated Object Detection}
Although centralized training remains a strong baseline, it assumes that all images can be pooled in a single repository. This assumption is often unrealistic in drone deployments, where data may be geographically distributed, bandwidth-limited, operationally sensitive, or subject to privacy and security constraints. FL addresses this limitation by allowing nodes to collaboratively train a shared model without transmitting raw images~\cite{mcmahan2017communication, yang2019federated}.

Prior work has already shown the promise of FL for visual perception. FedVision~\cite{liu2020fedvision} provided an early practical framework for federated visual object detection. However, drone-specific federated object detection remains considerably less explored than generic federated vision tasks. In one of the few domain-specific studies, Lu and Sun~\cite{lu2024development} developed a real-time UAV object detection system trained with FL and demonstrated that the resulting model can be deployed on a real drone while maintaining practical inference speed. More recently, Lu and Sun~\cite{lu2025improving} addressed non-IID data in federated UAV object detection by introducing model contrastive loss together with focal loss, showing clear improvements over standard FedAvg-based training. Taken together, these studies suggest that FL is a promising approach for drone-based object detection, while also highlighting that robustness to non-IID data and deployment efficiency remain central challenges.

Compared with this literature, our work focuses on a drone-based object-detection scenario using KIIT-MiTA and evaluates centralized, single-node, and federated training under a common experimental pipeline. In this sense, the contribution is not a new detector architecture or a new FL optimizer, but a controlled study of whether federated training can preserve competitive aerial-detection performance in a realistic drone-imagery setting while avoiding raw-data centralization.

\subsection{Problem Description}
\label{sec:prob-description}

The object detection task consists of jointly performing \emph{object localization} and \emph{classification} within an input image. Formally, given an input image \( x \in \mathcal{X} \), the goal is to identify a variable number of objects present in the image, predict their corresponding class labels, and localize them via bounding boxes.

\paragraph{Output Structure.}
Each annotation associated with an image is denoted as \( y \in \mathcal{Y} \) and consists of a set of tuples:
\[
    y = \{ (b_j, c_j) \}_{j=1}^{M},
\]
where \( M \) is the (variable) number of objects in the image. For each object:
\begin{itemize}
    \item \( b_j \in \mathbb{R}^4 \) represents the bounding box coordinates of the \( j \)-th object, typically encoded as either the center coordinates along with width and height, or the top-left and bottom-right corners of the box.
    \item \( c_j \in \{1, \dots, C\} \) denotes the corresponding class label from a predefined set of \( C \) object categories.
\end{itemize}

\paragraph{Comparison with Standard Classification.} Unlike standard image classification tasks where each image corresponds to a single label, object detection requires \emph{structured prediction} with a variable-length output for each input sample. This introduces additional complexity in model design, training, and inference, as the model must simultaneously:
\begin{enumerate}
    \item Predict the number of objects \( M \) present in the image.
    \item Regress the bounding box coordinates \( \{ b_j \} \).
    \item Classify each object into one of the \( C \) categories via \( \{ c_j \} \).
\end{enumerate}

\paragraph{Objective.} Given a dataset of \( N \) training samples \( \{ (x_i, y_i) \}_{i=1}^{N} \), the objective in object detection is to learn a mapping:
\[
    f_\theta: \mathcal{X} \to \mathcal{Y},
\]
parameterized by \( \theta \), that accurately predicts the set of bounding boxes and class labels for each image. The following formulation is intended as a generic object-detection objective. The exact loss terms depend on the YOLO version used in each experiment.

\paragraph{Loss Computation.} The loss function in object detection typically combines a localization loss, e.g., smooth $L_1$ or Intersection over Union (IoU) based loss, for bounding box regression and a classification loss (e.g., cross-entropy loss) for category prediction. Let \( \mathcal{L}_{\text{loc}} \) denote the localization loss and \( \mathcal{L}_{\text{cls}} \) denote the classification loss; then the total loss can be expressed as:
\[
    \mathcal{L}(\theta) = \lambda_{\text{loc}} \cdot \mathcal{L}_{\text{loc}}(\theta) + \lambda_{\text{cls}} \cdot \mathcal{L}_{\text{cls}}(\theta),
\]
where \( \lambda_{\text{loc}} \) and \( \lambda_{\text{cls}} \) are hyperparameters to balance the contributions of the localization and classification terms.

\section{ML Privacy-Preserving Solution}

In this section, we provide a detailed explanation of the privacy-preserving ML solution, discuss privacy and regulatory limitations, and introduce FL.

\subsection{The ML Approach}

In the centralized (classical) ML paradigm, object detection is formulated as a supervised learning task. Given a labeled dataset $\mathcal{D} = \{(x_i, y_i)\}_{i=1}^N$, where $x_i \in \mathcal{X}$ represents the $i$-th input image and $y_i \in \mathcal{Y}$ denotes its corresponding ground truth annotations (bounding boxes and class labels), the objective is to learn a mapping function $f_\theta: \mathcal{X} \rightarrow \mathcal{Y}$ parameterized by $\theta \in \mathbb{R}^d$, such that $f_\theta(x_i)$ closely approximates $y_i$ for all $i$.

The training process involves minimizing an empirical risk over the dataset $\mathcal{D}$:

\begin{equation}
\min_{\theta \in \mathbb{R}^d} \mathcal{L}(\theta) = \frac{1}{N} \sum_{i=1}^N \ell(f_\theta(x_i), y_i),
\end{equation}

where $\ell: \mathcal{Y} \times \mathcal{Y} \rightarrow \mathbb{R}_{\geq 0}$ is a task-specific loss function that quantifies the discrepancy between the predicted and ground-truth annotations. In object detection, this loss typically comprises two main components:

\begin{equation}
\ell(f_\theta(x), y) = \ell_{\text{cls}}(f_{\theta}^{\text{cls}}(x), y^{\text{cls}}) + \lambda \cdot \ell_{\text{loc}}(f_{\theta}^{\text{loc}}(x), y^{\text{loc}}),
\end{equation}

where $\ell_{\text{cls}}$ is the classification loss (e.g., cross-entropy), $\ell_{\text{loc}}$ is the localization loss, and $\lambda > 0$ is a hyperparameter controlling the trade-off between the two objectives. Here, $f_\theta^{\text{cls}}(x)$ and $f_\theta^{\text{loc}}(x)$ denote the classification and localization outputs of the model, respectively.

Training is typically performed using variants of stochastic gradient descent (SGD) or adaptive gradient methods such as Adam, where the parameters are updated iteratively as:

\begin{equation}
\theta_{t+1} = \theta_t - \eta_t \nabla_\theta \mathcal{L}(\theta_t),
\end{equation}

with $\eta_t$ denoting the learning rate at iteration $t$. This iterative optimization continues until convergence or early stopping criteria are met based on validation performance.

While effective in controlled settings, this classical approach becomes impractical in real-world applications involving data privacy constraints, regulatory barriers, or high communication costs. For example, in our object detection setting, transferring visual data from edge devices to a central server is both privacy-sensitive and bandwidth-intensive.
FL addresses these limitations by enabling collaborative training without requiring raw data to leave its source.

\subsection{Introduction to FL}
\label{sec:fedlearning-overview}

In FL~\cite{mcmahan2017communication}, each node \(k\) maintains a local model \(f_{\theta_k}\) and periodically transmits model parameters or gradients to a central server (the \emph{aggregator}).  The server combines these updates into a global model and redistributes it to each node. This cycle repeats until convergence.
Although FL preserves data privacy and allows edge devices to train with their data, it also introduces several challenges. The most notable one is related to handling non‐IID data across nodes, which can yield divergent local updates and degrade global accuracy. This problem is also referred to in the literature as \textit{data drift} or \textit{concept drift}~\cite{lu2024federated}.

%FL presents a framework for decentralized model training, allowing local data to remain on node devices while only model updates are communicated. 

In the context of computer vision, FL has been successfully explored for tasks such as image classification~\cite{caldas2018leaf, li2020federated} and medical image analysis~\cite{sheller2020federated}. However, applying FL to object detection remains more challenging due to the typically larger model sizes and more complex loss functions (involving localization and classification simultaneously)~\cite{zhu2021federated}.

In this paper, we focus on a realistic FL setting in which each node retains a disjoint, private subset of the data, simulating the practical scenario where data naturally reside across different edge devices or drones in a federated environment (see Section~\ref{sec:nodes_creation}).

\subsubsection{FL Paradigms}
\label{sec:fedlearning-paradigms}

Depending on the data distribution among nodes, there are two FL paradigms:
\begin{itemize}
    \item \textbf{HFL}~\cite{yang2019federated}: In this paradigm, all nodes share the same feature space but hold different samples (rows). \emph{Example}: Multiple drones collect images using similar onboard cameras, but each drone captures different aerial scenes, operational areas, and target instances under varying altitudes, viewpoints, and environmental conditions.
  
    \item \textbf{Vertical FL (VFL)}: Nodes have complementary feature subsets for the same samples (shared index set). \emph{Example}: One drone subsystem captures visual data from onboard cameras, while another records corresponding thermal imagery, GPS coordinates, or altitude measurements for the same aerial scene. Together, they provide different feature views of the same observation.
\end{itemize}

In this work, we focus on HFL. After training is complete in an HFL setup, the resulting global model is typically shared with all participating nodes. This allows each node to download the trained model and subsequently perform inference locally and independently, without requiring further interaction or data exchange.

\subsubsection{HFL}
\label{sec:fedlearning-hfl}

Under HFL, each node \(k\) has a local dataset:
\[
\mathcal{D}_k
= \bigl\{(\mathbf{x}_i^k,\,\mathbf{y}_i^k)\bigr\}_{i=1}^{N_k},
\]
where all \(\mathbf{x}_i^k\in\mathbb{R}^d\) share the same feature dimension \(d\), but the number of samples \(N_k\) can differ.  Training proceeds as follows:
\begin{enumerate}
  \item \textbf{Local update:} Each node \(k\) performs local optimization on its empirical risk, typically for a fixed number of local epochs.
\[
\mathcal{J}_k(\theta)
= \frac{1}{N_k}
  \sum_{i=1}^{N_k}
    \mathcal{L}\bigl(f_\theta(\mathbf{x}_i^k),\,\mathbf{y}_i^k\bigr)
\]
  \item \textbf{Aggregation:} Nodes send their updated parameters \(\theta_k\) to the server.
  \item \textbf{Global update:} The server aggregates the local model parameters \(\{\theta_k\}\) into updated global parameters \(\theta\).
  \item \textbf{Broadcast:} The server distributes \(\theta\) back to all nodes and the process starts again.
\end{enumerate}

The training objective is to minimize a global loss:
\begin{equation}
\min_{\theta} \sum_{k=1}^{K} \frac{N_k}{N} \mathcal{J}_k(\theta),
\end{equation}
where the total number of samples $N = \sum_{k=1}^{K} N_k$, and $\mathcal{J}_k(\theta)$ is the local loss evaluated on $\mathcal{D}_k$.

HFL is particularly suited for scenarios where different data holders have datasets with the same feature space but different samples. This setup aligns naturally with many real-world object detection applications where similar types of imagery are collected independently by different devices. In these cases, HFL enables collaborative model training that benefits from a greater diversity of training examples while respecting data locality and privacy constraints.

\section{Centralized Dataset and Preprocessing}

In this section, we describe the dataset, outline the preprocessing steps, and present the centralized architecture.

\subsection{Description of the Dataset} \label{sec:descrip_data}
The dataset employed in this study is the KIIT-MiTA dataset, obtained from Kaggle~\cite{chakrabarty2025kiitmita}. KIIT-MiTA is a drone-imagery dataset designed for military object detection and recognition, and was created to support the development and evaluation of computer vision models for aerial surveillance scenarios. The data acquisition is based on aerial imagery captured from drone platforms, reflecting realistic operational conditions in which targets appear at different scales, viewpoints, and scene complexities. This makes the dataset particularly suitable for studying object detection in drone-based perception systems.

\begin{figure}[h!]
    \centering
    \includegraphics[width=1.0\linewidth]{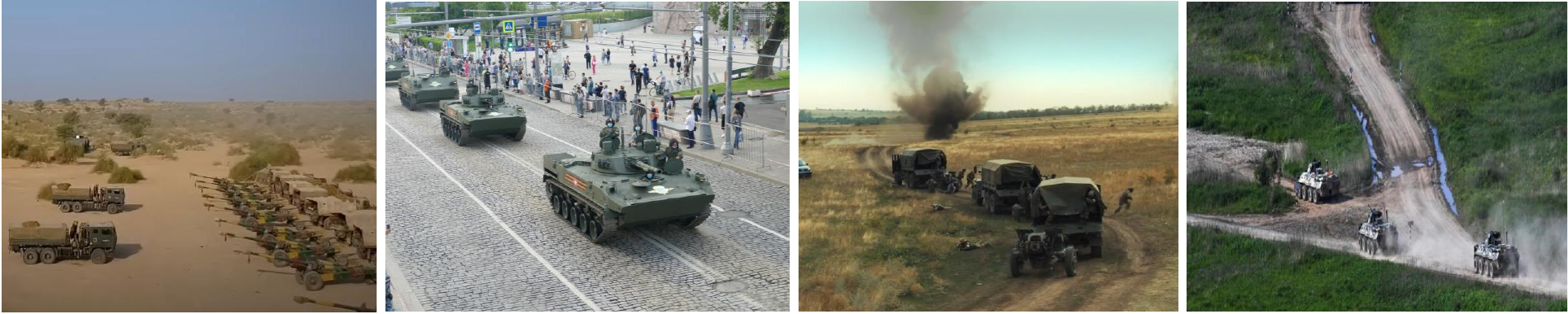}
    \caption{Example images from the KIIT-MiTA dataset~\cite{chakrabarty2025kiitmita} showing some military target classes in an aerial drone scene.}
    \label{fig:KIITMITAExample}
\end{figure}

The dataset consists of high-resolution RGB drone images annotated with bounding boxes corresponding to seven military-relevant classes as illustrated in Figure~\ref{fig:KIITMITAExample}. These annotations enable supervised training of deep learning models by providing both localization and semantic class information for each target instance. The dataset is especially challenging due to the aerial viewpoint, the presence of small objects, variations in altitude and scale, cluttered backgrounds, and differences in environmental conditions. Such characteristics make KIIT-MiTA well-suited for evaluating object detection systems in distributed and safety-critical drone applications.

% # Poner imágenes que sean más similares a casos reales (proponer imagenes)

% \daniel{Figure~\ref{fig:KIITMITAExample} must be updated by Oleksii from the actual dataset. DONE.}
%\oleksii{Done.}

In our study, we utilize all annotated categories during training, resulting in a total of seven classes: artillery, missile, radar, multiple rocket launcher (M. R. Launcher), soldier, tank, and vehicle.

\subsection{Preprocessing of the Dataset}
\label{sec:preprocessing}

The preprocessing pipeline was adapted to the training workflow used in the Ultralytics YOLO framework~\cite{ultralytics_framework}. In particular, images and annotations were organized in the format required by the Ultralytics framework, and standard augmentation strategies were applied during training to improve robustness and generalization. The pipeline includes:

\begin{itemize}
    \item \textit{Image preparation:} Raw image files were loaded and resized to the input resolution required by the model during training.
    \item \textit{Annotation preparation:} Bounding box annotations and class labels were converted into the format expected by the Ultralytics training pipeline.
    \item \textit{Data augmentation:} Built-in Ultralytics YOLO augmentations~\cite{ultralytics_framework} were applied during training with the following settings: mosaic $=1.0$, mixup $=0.05$, copy-paste $=0.0$, horizontal flip $=0.5$, vertical flip $=0.0$, translation $=0.10$, scale $=0.50$, and HSV augmentation $(h=0.015, s=0.70, v=0.40)$. Rotation, shear, and perspective augmentation were disabled. These augmentation parameters were selected empirically after preliminary experiments with different configurations. The final setup was chosen because it provided the best validation performance while maintaining realistic transformations for drone imagery and avoiding augmentations that degraded detection accuracy.
\end{itemize}

The system architecture used for centralized training is shown in Figure~\ref{fig:cen-arch}. In our experiments, we used the official training, validation, and test partitions provided with the dataset. Under the centralized setting, all data from the training partition is assumed to be available at a central server for model optimization, while validation and test partitions are used for model selection and final evaluation, respectively.

\subsection{Centralized Architecture}
\label{sec:centr_arch}

\begin{figure}[!ht]
    \centering
    \includegraphics[scale=0.5]{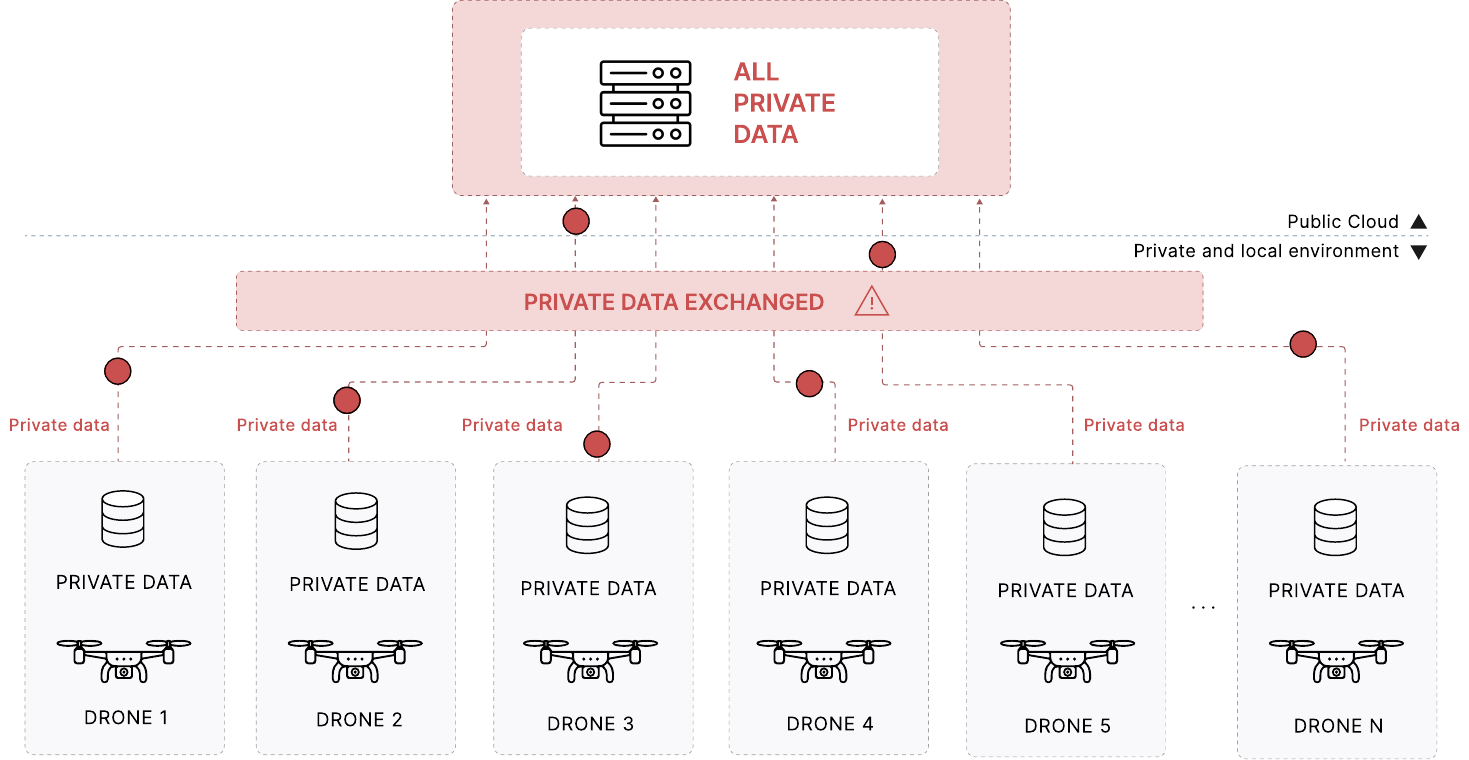}
    \caption{Proposed architecture for centralized training.}
    \label{fig:cen-arch}
\end{figure}

To establish a performance baseline and validate our training pipeline, we conducted a centralized object detection experiment (see Figure~\ref{fig:cen-arch}) using the Ultralytics framework~\cite{ultralytics_framework} and models from the YOLO family. This centralized setting assumes that all training data is available at a central server, allowing the selected detector to be optimized using the complete training partition of the dataset.

\section{Proposed Privacy-Preserving Solution through FL} \label{sec:fl-solution}

In this section, we present the FL architecture and the construction of the federated nodes used in our experimental setup.

\subsection{FL Architecture} 

\begin{figure}[!ht]
    \centering
    \includegraphics[scale=0.5]{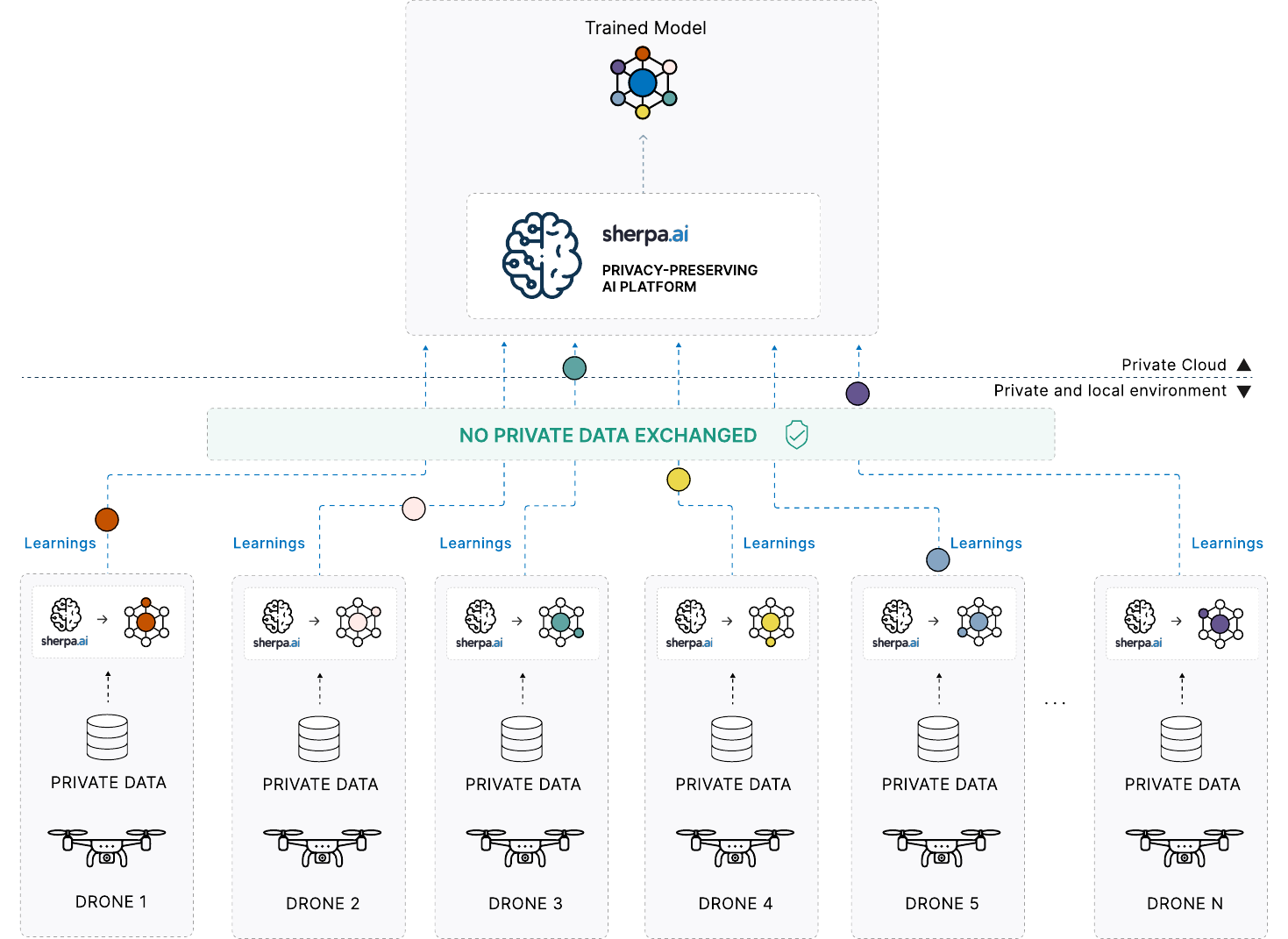}
    \caption{Proposed architecture for federated training.}
    \label{fig:fed-arch}
\end{figure}

We implement FL as depicted in Figure~\ref{fig:fed-arch}. A global model is initialized and distributed to all nodes, each of which performs local training using its private data. After a fixed number of local epochs, model updates (weights) are communicated to the Platform, which performs aggregation via Federated Averaging (FedAvg)~\cite{mcmahan2017communication} and then distributes the updated model parameters back to the nodes. This iterative process continues over multiple communication rounds until convergence. Crucially, no raw data are transmitted during training, which preserves data locality and reduces exposure to potential data leakage compared with centralized data collection. Thus, each node receives the updated federated model and can subsequently perform inference locally, if needed.

When applying this framework to object detection, local training involves optimizing a compound loss function encompassing classification, objectness, and bounding box regression. Given the model's computational demands, the efficiency of local computation and communication overhead becomes critical.

%\george{Considering the complexity of ultralytics, I think we should mention what we choose to aggregate and what we keep local during FL.}

\subsection{Creation of Nodes} \label{sec:nodes_creation}

To simulate a realistic federated setting in which each node retains its own local data, the training split of the dataset was partitioned across \(K=4\) nodes using a Dirichlet-based protocol~\cite{jimenez2024fedartml}. This partitioning strategy creates non-IID data distributions across nodes, as commonly observed in practical federated environments where participants may capture different target types, scene compositions, and operational conditions.

The partitioning process was controlled by a concentration parameter \(\alpha=1\), which induces a moderate level of non-IID data among nodes. As a result, each node receives a different proportion of training examples per class while preserving the overall dataset structure. This setup provides a realistic and reproducible benchmark for evaluating federated object detection under non-IID data conditions.

It is important to note that, in the case of object detection, non-IID partitioning is more complex than in standard image classification. Each image may contain multiple object instances and, potentially, multiple object classes. Therefore, assigning an image to a single class for partitioning purposes can only provide an approximate description of the data heterogeneity. In our partitioning protocol, each training image was assigned a dominant class, defined as the class with the largest number of annotated bounding boxes in that image. The Dirichlet-based split was then applied at the image level using these dominant-class labels. As a result, each image and all of its associated bounding-box annotations were assigned to a single node, avoiding any leakage of image-level information across nodes.

\begin{figure}[h]
\centering
  \includegraphics[scale=0.45]{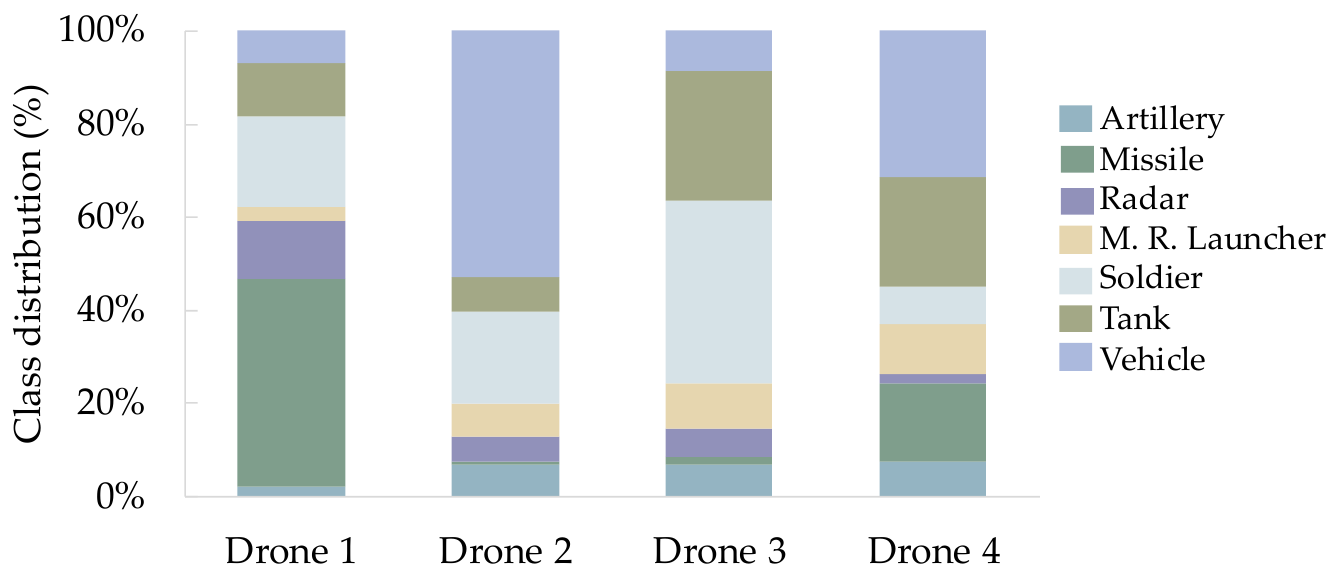}
    \caption{Dominant-class distribution across drones for the non-IID partition used. Each stacked bar shows the proportion of images assigned to each class according to the most frequent object class in the image.}
  \label{fig:distro_kiit}
\end{figure}

Figure~\ref{fig:distro_kiit} shows the dominant-class distributions obtained across the four drones after applying the Dirichlet-based partitioning strategy. The resulting distributions exhibit heterogeneous class proportions across drones, reflecting the non-IID setting used in the federated experiments. For images containing multiple object classes, each image was assigned to the class appearing most frequently in that image.

% \daniel{The previous two paragraphs must be checked once the final results are obtained by Oleksii.}
% \oleksii{Seems correct.}

\section{Experiments} \label{sec:experiments}

In this section, we describe the experimental setup and the training configurations considered in this work, and present the main results for drone-based object detection.

We evaluate model performance across three training scenarios to assess the benefits of FL for the KIIT-MiTA drone dataset:

\begin{enumerate}
    \item \textbf{Centralized}: All training data is assumed to be available at a central server, and the model is trained using the complete training partition of the dataset. Evaluation is then performed on the official test set.
    
    \item \textbf{Federated}: The training partition of the dataset is distributed across multiple nodes according to the Dirichlet-based protocol described in Section~\ref{sec:nodes_creation}. Each node performs local training on its own private data, and the local updates are aggregated by the central server over successive communication rounds. The resulting global model is then evaluated on the official test set.
    
    \item \textbf{Single-drone}: Each node trains its own model independently using only its local partition of the training data, without any collaboration or parameter aggregation. Each locally trained model is evaluated on the official test set. The Single-drone results reported in the tables correspond to the mean performance across the four nodes.
\end{enumerate}

These three scenarios enable a direct comparison between full data centralization, fully isolated local training, and collaborative training under privacy-preserving constraints. As such, the experiments quantify the extent to which FL provides an effective compromise between avoiding raw data sharing and still benefiting from the distributed knowledge available across nodes.

\subsection{Evaluation Metrics}
\begin{figure}[htb]
  \centering
  \includegraphics[width=1.0\textwidth]{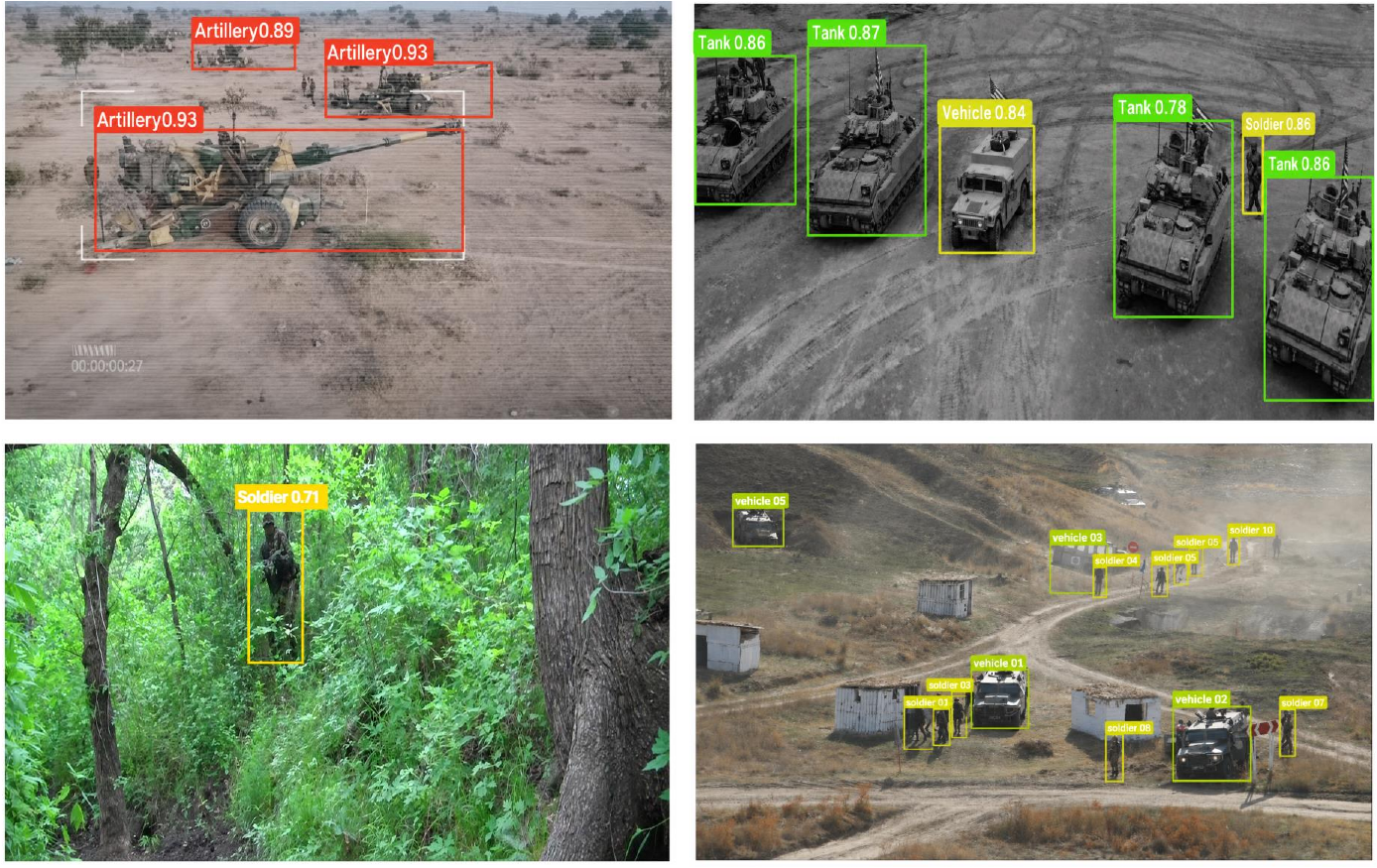}
    \caption{Qualitative examples of model predictions on test images. Detected objects are shown with their predicted bounding boxes, class labels, and confidence scores.}
  \label{fig:output}
\end{figure}

Figure~\ref{fig:output} shows qualitative examples of the detections produced by the trained model on representative test images. The visualizations illustrate the two main outputs of the object detection task: object localization through bounding boxes and class prediction through the corresponding labels and confidence scores.

While these qualitative results provide an intuitive view of the model predictions, a quantitative evaluation is required to compare the different training scenarios. To this end, we consider both predictive and efficiency metrics. For predictive performance, we employ two standard measures: \textit{mean Average Precision} at an IoU threshold of 0.50 (mAP@0.50) and mean Average Precision averaged over IoU thresholds from 0.50 to 0.95 (mAP@0.50:0.95), in line with COCO-style object detection evaluation~\cite{lin2014microsoft}. These metrics are also reported by the Ultralytics validation pipeline~\cite{ultralytics_framework}. In addition, to assess computational efficiency, we report Frames Per Second (FPS) and Giga Floating Point Operations (GFLOPs). In the following, we detail the calculation of these evaluation metrics.

\paragraph{Precision and Recall.} Precision measures the proportion of correctly predicted positive detections among all positive detections made by the model. Let $TP$ denote the number of true positives and $FP$ the number of false positives. Then, precision is defined as:

\begin{equation}
\text{Precision} = \frac{TP}{TP + FP}
\end{equation}

In the context of object detection, a predicted bounding box is considered a \textit{true positive} if its IoU with a ground truth box is greater than a given threshold (i.e., $\tau = 0.50$), and the predicted class matches the ground truth class. Otherwise, it is a \textit{false positive}.

Recall, on the other hand, measures the proportion of correctly predicted positive detections among all actual positives in the ground truth. Let $FN$ denote the number of false negatives, i.e., ground truth objects not detected by the model. Recall is then defined as:

\begin{equation}
\text{Recall} = \frac{TP}{TP + FN}
\end{equation}

A false negative corresponds to a ground truth object for which no predicted bounding box satisfies the IoU threshold or class match. High recall indicates that most actual objects are being detected, whereas low recall suggests that the model is missing a significant number of objects present in the scene.

\paragraph{AP@0.50.} For a fixed IoU threshold $\tau = 0.50$, AP for a given class $c$ is computed as the area under the precision-recall curve. This curve is generated by varying the confidence threshold $\theta$ used to accept a detection. At each $\theta$, the set of predictions is filtered, and true positives (TP), false positives (FP), and recall values are recalculated accordingly.

Let $\mathcal{P}_c(\theta)$ and $\mathcal{R}_c(\theta)$ denote the precision and recall at threshold $\theta$ for class $c$, respectively. Then, the AP at $\tau = 0.50$ for class $c$ is given by:

\begin{equation}
AP_{\tau=0.50}^c = \int_0^1 \mathcal{P}_c(r) \, dr,
\end{equation}
where the integration is performed with respect to recall. 

In practice, this integral is approximated using a discrete set of recall points $\{r_1, r_2, \dots, r_N\}$, leading to:

\begin{equation}
AP_{\tau=0.50}^c \approx \sum_{n=1}^N (\mathcal{R}_c(r_n) - \mathcal{R}_c(r_{n-1})) \cdot \mathcal{P}_c(r_n)
\end{equation}

where the set of $(\mathcal{P}_c(r_n), \mathcal{R}_c(r_n))$ pairs is obtained by sorting the detections by confidence and evaluating the precision and recall at each point.

\paragraph{mAP@0.50} To obtain the overall performance across classes, the mAP at IoU threshold 0.50 is computed as the mean of APs across all $C$ object classes:

\begin{equation}
\text{mAP@0.50} = \frac{1}{C} \sum_{c=1}^C AP_{\tau=0.50}^c
\end{equation}

This metric, denoted as mAP@0.50, is commonly used in object detection benchmarks and reflects the average precision when an object is considered correctly detected if the IoU exceeds 0.50.

\paragraph{mAP@0.50:0.95} The COCO evaluation protocol and Ultralytics introduce an even more stringent metric, denoted as mAP@0.50:0.95, which averages the AP also over multiple IoU thresholds to better capture both localization and classification accuracy. Specifically, it computes the mean of APs across 10 IoU thresholds from 0.50 to 0.95 in increments of 0.05:

\begin{equation}
\text{mAP@0.50:0.95} = \frac{1}{10C} \sum_{\tau \in \{0.50, 0.55, \ldots, 0.95\}} \sum_{c=1}^C AP_{\tau}^c
\end{equation}

This formulation penalizes poor localization more heavily and provides a more comprehensive assessment of detector performance, especially for models evaluated on diverse datasets with varied object scales and occlusion patterns.

\paragraph{FPS.} It quantifies the inference throughput of the object detection model, representing the number of image frames the model can process per second during inference. It is defined as:
\[
\mathrm{FPS} = \frac{N}{T}
\]
where \( N \) is the number of images processed, and \( T \) is the total time (in seconds) taken to process those images. A higher FPS value indicates faster inference speed and better real-time performance, which is critical in applications such as object detection with drones. FPS is typically measured using single-stream inference on a fixed hardware platform, often without including pre- and post-processing steps unless otherwise stated.

\paragraph{GFLOPs.} It measures the theoretical computational complexity of the model in terms of billions of floating-point operations required per forward pass during inference. For a given model, GFLOPs measures the number of floating-point operations per forward pass and is defined as:
\[
\mathrm{GFLOPs} = \frac{\mathrm{FLOPs}}{10^9}
\]
FLOPs are computed as the sum of all multiply–accumulate operations (MACs), which are then typically multiplied by two (to account for one multiplication and one addition). GFLOPs is a hardware-agnostic indicator of model complexity and does not reflect actual wall-clock speed. It is particularly useful for comparing computational demands across models independently of implementation or hardware specifics. For this metric, lower values indicate better computational efficiency, provided that the model maintains acceptable accuracy.

\subsection{Reproducibility Details}
\label{sec:reproducibility-details}

The experiments were carried out using two hardware configurations:

\begin{enumerate}
    \item The centralized and single-drone experiments, as well as inference and model validation, were conducted on a workstation equipped with a 13th Gen Intel(R) Core(TM) i7-13700K CPU and an NVIDIA RTX A4000 GPU. The system included 93~GiB of RAM and 8~GiB of swap memory. GPU acceleration was provided by an NVIDIA RTX A4000 with 16,376~MiB of dedicated memory. The GPU driver version was 560.35.05, the CUDA version was 12.6, the operating system was Ubuntu 24.04, and the Python version was 3.11.

    \item The federated experiments were executed on four AWS EC2 g4dn instances: three \texttt{g4dn.xlarge} instances and one \texttt{g4dn.2xlarge} instance. The \texttt{g4dn.xlarge} instances provide 4 vCPUs and 16~GiB of memory, while the \texttt{g4dn.2xlarge} instance provides 8 vCPUs and 32~GiB of memory. Both instance types belong to the GPU-accelerated G4dn family and are equipped with NVIDIA T4 GPUs. The Amazon Machine Image used was Ubuntu 22.04 and Python 3.11.
\end{enumerate}

For a fair comparison of inference efficiency, FPS values were measured using the same validation hardware across all trained models.

The federated training process follows the FedAvg algorithm~\cite{mcmahan2017communication}. In each communication round, every participating node performs 10 local training epochs before sending its updated model parameters to the server for aggregation. The training is conducted for a total of 10 communication rounds. All models are trained using the \sherpa FL platform.

Unless otherwise stated, the same hyperparameter configuration is used across all experimental settings to ensure a fair comparison. In particular, all models are trained with an input image size of $640 \times 640$ pixels and a batch size of 16, while the optimizer and learning rate follow the default configuration of the corresponding Ultralytics model. For the Centralized and Single-drone baselines, models are trained for 100 epochs, matching the overall training budget used in the federated setting. The evaluated detectors were YOLOv5 nano, YOLOv8 nano, YOLO11 nano, and YOLO26 nano, implemented through the Ultralytics framework using the corresponding nano checkpoints.

Each experimental configuration was repeated over 10 independent runs using different random seeds. The accuracy values reported in Tables~\ref{tab:results} and~\ref{tab:yolo26_per_class_results} correspond to the mean and standard deviation across these runs. For the Single-drone setting, the metric was first averaged across the four independently trained nodes in each run.

% \daniel{The previous hyperparameters must be updated for the ones used b Oleksii}
% \oleksii{Done.}

% \daniel{The previous detail of Section~\ref{sec:reproducibility-details} should be updated by the ones used in the experiments by Oleksii.}
% \oleksii{Done}

\subsection{Results} \label{sec:results}

This section reports the experimental results obtained across the three training scenarios considered in this work: Single-drone, Centralized, and Federated. The analysis first compares the global detection performance of the evaluated YOLO nano models, then examines their inference speed and computational complexity to assess suitability for drone-based deployment. We further analyze the per-class performance of the best-performing model and conclude by evaluating the robustness of FL under bandwidth-constrained conditions representative of wireless and edge environments.

\subsubsection{Global Detection Performance}

Table~\ref{tab:results} presents the evaluation results of all experiments on the test images introduced in Section~\ref{sec:descrip_data}. Overall, the performance achieved in our experiments is consistent with previous work on object detection using this dataset~\cite{chakrabarty2025drones}. Notably, among the evaluated models, YOLO26 nano achieves the strongest performance and compares favorably with previously reported results for this dataset under similar evaluation metrics.

\begin{table}[h]
  \scriptsize
  \setlength{\tabcolsep}{4pt}
  \renewcommand{\arraystretch}{1.25}
  \centering
  \begin{tabularx}{\textwidth}{c|c|YYY}
    \toprule
    \textbf{Model} & \textbf{Metric} & \textbf{Single-drone} & \textbf{Centralized} & \textbf{Federated} \\
    \midrule
    \multirow{5}{*}{YOLOv5 nano}
      & mAP@0.50        & 0.4229 (0.0676) & 0.6804 (0.0242) & 0.6575 (0.0199) \\
      & mAP@0.50:0.95   & 0.2444 (0.0439) & 0.4292 (0.0265) & 0.4118 (0.0151) \\
      % & FPS (CPU)       & 30.8 & 30.8 & 30.8 \\
      & FPS (GPU)       & 105.7 & 106.0 & 102.9 \\
      & GFLOPs          & 7.8 & 7.8 & 7.8 \\
    \midrule
    \multirow{5}{*}{YOLOv8 nano}
      & mAP@0.50        & 0.4234 (0.0605) & 0.6939 (0.0159) & 0.6839 (0.0150) \\
      & mAP@0.50:0.95   & 0.2487 (0.0403) & 0.4568 (0.0123) & 0.4360 (0.0074) \\
      % & FPS (CPU)       & 29.6 & 29.6 & \\
      & FPS (GPU)       & 104.5 & 105.6 & 104.5 \\
      & GFLOPs          & 8.9 & 8.9 & 8.9 \\
    \midrule
    \multirow{5}{*}{YOLO11 nano}
      & mAP@0.50        & 0.4225 (0.0499) & 0.7046 (0.0157) & 0.6984 (0.0186) \\
      & mAP@0.50:0.95   & 0.2502 (0.0324) & 0.4618 (0.0144) & 0.4633 (0.0141) \\
      % & FPS (CPU)       & 26.7 & 26.7 & 26.7 \\
      & FPS (GPU)       & 101.4 & 101.7 & 101.4 \\
      & GFLOPs          & 6.6 & 6.6 & 6.6 \\
    \midrule
    \multirow{5}{*}{YOLO26 nano}
      & mAP@0.50        & 0.4668 (0.0648) & 0.7322 (0.0156) & 0.7137 (0.0153) \\
      & mAP@0.50:0.95   & 0.2798 (0.0436) & 0.4959 (0.0176) & 0.4695 (0.0149) \\
      % & FPS (CPU)       & 26.1 & 26.1 & 26.1 \\
      & FPS (GPU)       & 98.6 & 99.1 & 98.6 \\
      & GFLOPs          & 6.1 & 6.1 & 6.1 \\
    \bottomrule
  \end{tabularx}
  \vspace{4mm}
  \caption{Metric comparison across models for Single-drone, Centralized, and Federated scenarios. mAP metrics are reported as mean values with standard deviations (STD) over 10 independent runs with different random seeds. FPS and GFLOPs are reported as model-level values.}
  \label{tab:results}
\end{table}

The improvement of Federated training over the Single-drone baseline is substantial across all models. For example, in terms of mAP@0.50, the Federated setting improves over Single-drone training by 0.2346 for YOLOv5 nano, 0.2605 for YOLOv8 nano, 0.2759 for YOLO11 nano, and 0.2469 for YOLO26 nano. Similar gains are observed for the stricter mAP@0.50:0.95 metric, where the corresponding improvements are 0.1674, 0.1873, 0.2131, and 0.1897, respectively. These results confirm that isolated local training is not sufficient to fully capture the diversity of the distributed drone imagery, whereas Federated training allows the global model to benefit from knowledge aggregated across nodes.

\begin{figure}[h]
\centering
  \includegraphics[scale=0.5]{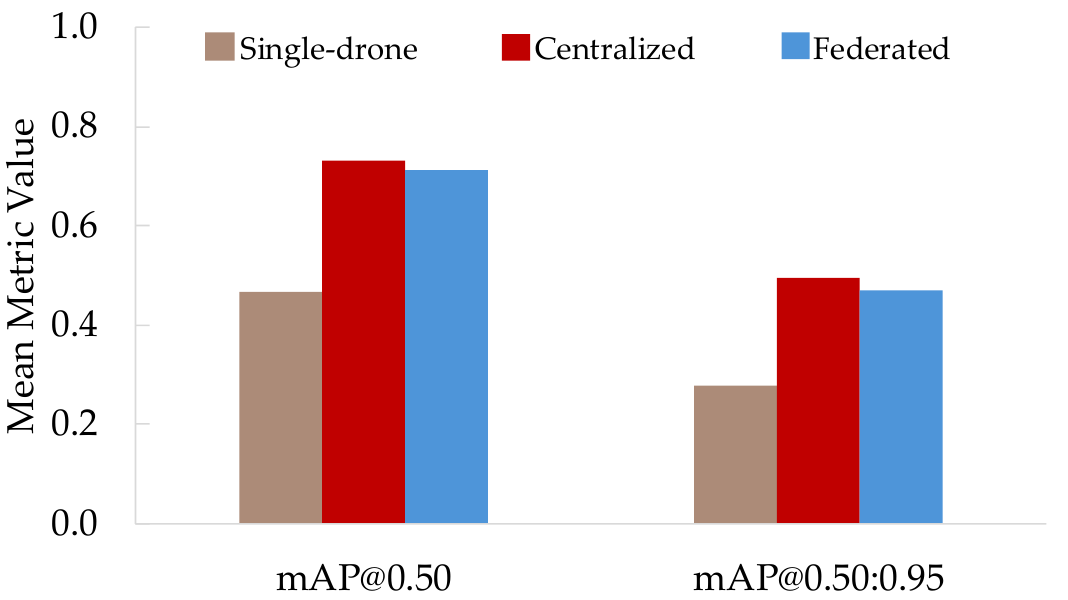}
  \caption{Mean metric values comparing Single-drone, Centralized, and Federated training settings for the best model (YOLO26 nano).}
  \label{fig:results}
\end{figure}

Among the evaluated models, YOLO26 nano achieves the best overall accuracy. In the Federated setting, it obtains the highest mAP@0.50, with a value of 0.7137, and the highest mAP@0.50:0.95, with a value of 0.4695. Compared with the Single-drone baseline, this represents an absolute gain of 0.2469 in mAP@0.50 and 0.1897 in mAP@0.50:0.95, corresponding to relative improvements of approximately 52.89\% and 67.80\%, respectively. Moreover, the Federated YOLO26 nano model remains close to its Centralized counterpart. This shows that the Federated model retains most of the accuracy of centralized training while preserving data locality.

Figure~\ref{fig:results} provides a focused visual comparison for YOLO26 nano, the best-performing model in Table~\ref{tab:results}. The figure highlights the same trend observed in the full table: Federated training clearly outperforms the Single-drone baseline and approaches the Centralized baseline for both evaluation metrics. This result supports the use of FL as a practical alternative to centralized training for drone-based object detection, since it achieves strong detection performance without requiring the transfer of raw data among nodes.

\subsubsection{Inference Speed and Complexity}

Inference speed and computational complexity are also important for drone-based deployment, where models must operate under latency and hardware constraints. As shown in Table~\ref{tab:results}, all evaluated nano models achieve high GPU inference throughput, with FPS values ranging from 98.6 to 106.0 across the three training scenarios. The FPS values remain nearly identical for Single-drone, Centralized, and Federated training within each model, indicating that the training strategy does not introduce additional inference-time overhead.

In terms of computational complexity, YOLO26 nano is the most efficient model among those evaluated, requiring only 6.1 GFLOPs, compared with 6.6 GFLOPs for YOLO11 nano, 7.8 GFLOPs for YOLOv5 nano, and 8.9 GFLOPs for YOLOv8 nano. Combined with its best overall detection accuracy, this makes YOLO26 nano the most favorable in terms of accuracy and efficiency in our experiments. In particular, the Federated YOLO26 nano model achieves 98.6 FPS with 6.1 GFLOPs, showing that the proposed FL approach preserves real-time inference capability while avoiding raw data centralization.

\subsubsection{Per-class Detection Performance}

Table~\ref{tab:yolo26_per_class_results} provides a per-class analysis of the best-performing model, YOLO26 nano. The results show that Federated training consistently improves over the Single-drone baseline for all classes and both mAP metrics, confirming that collaborative learning benefits not only the global mAP but also the detection performance at the class level.

\begin{table}[h]
  \scriptsize
  \setlength{\tabcolsep}{4pt}
  \renewcommand{\arraystretch}{1.25}
  \centering
  \begin{tabularx}{\textwidth}{c|c|YYY}
    \toprule
    \textbf{Class} & \textbf{Metric} & \textbf{Single-drone} & \textbf{Centralized} & \textbf{Federated} \\
    \midrule
    \multirow{2}{*}{Artillery}
      & mAP@0.50        & 0.4400 (0.1525) & 0.7673 (0.0462) & 0.7748 (0.0354) \\
      & mAP@0.50:0.95   & 0.2201 (0.0878) & 0.4730 (0.0398) & 0.4539 (0.0243) \\
    \midrule
    \multirow{2}{*}{Missile}
      & mAP@0.50        & 0.3165 (0.0897) & 0.7275 (0.0364) & 0.6918 (0.0477) \\
      & mAP@0.50:0.95   & 0.2048 (0.0782) & 0.5737 (0.0377) & 0.5302 (0.0401) \\
    \midrule
    \multirow{2}{*}{Radar}
      & mAP@0.50        & 0.3592 (0.1398) & 0.6811 (0.0346) & 0.6031 (0.0269) \\
      & mAP@0.50:0.95   & 0.1744 (0.0737) & 0.3880 (0.0341) & 0.3205 (0.0238) \\
    \midrule
    \multirow{2}{*}{M. R. Launcher}
      & mAP@0.50        & 0.5258 (0.2530) & 0.8402 (0.0224) & 0.7677 (0.0391) \\
      & mAP@0.50:0.95   & 0.3864 (0.1959) & 0.6693 (0.0203) & 0.5933 (0.0422) \\
    \midrule
    \multirow{2}{*}{Soldier}
      & mAP@0.50        & 0.4927 (0.0594) & 0.6135 (0.0392) & 0.6515 (0.0202) \\
      & mAP@0.50:0.95   & 0.2588 (0.0443) & 0.3412 (0.0227) & 0.3670 (0.0188) \\
    \midrule
    \multirow{2}{*}{Tank}
      & mAP@0.50        & 0.6420 (0.1075) & 0.8297 (0.0246) & 0.8350 (0.0193) \\
      & mAP@0.50:0.95   & 0.4112 (0.0878) & 0.5837 (0.0224) & 0.5753 (0.0194) \\
    \midrule
    \multirow{2}{*}{Vehicle}
      & mAP@0.50        & 0.4916 (0.0895) & 0.6663 (0.0274) & 0.6718 (0.0253) \\
      & mAP@0.50:0.95   & 0.3028 (0.0634) & 0.4422 (0.0218) & 0.4464 (0.0198) \\
    \bottomrule
  \end{tabularx}
  \vspace{4mm}
  \caption{Per-class mAP comparison for YOLO26 nano across Single-drone, Centralized, and Federated scenarios. Metrics are reported as mean values with standard deviations (STD) over 10 independent runs with different random seeds.}
  \label{tab:yolo26_per_class_results}
\end{table}

\begin{figure}[h]
\centering
  \includegraphics[scale=0.5]{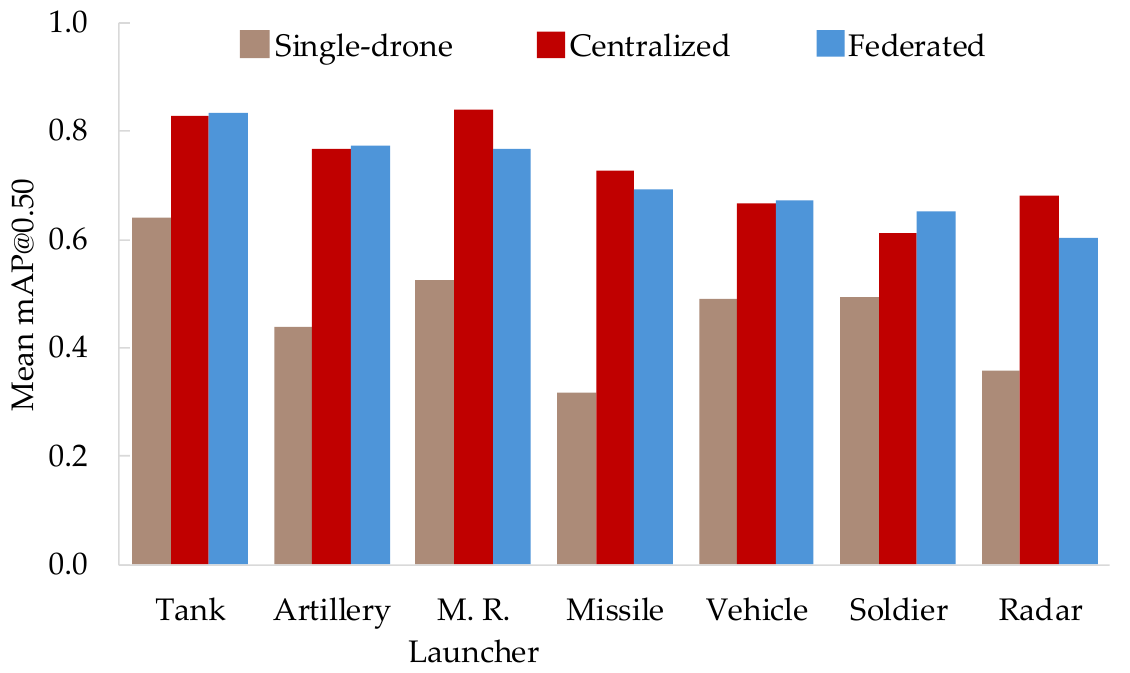}
  \caption{Per-class mean mAP@0.50 values comparing Single-drone, Centralized, and Federated training settings for the best model (YOLO26 nano).}
  \label{fig:per_class_results}
\end{figure}

The largest improvements are observed for classes such as Missile, Radar, Artillery, and Vehicle, where the Single-drone setting is limited by the reduced diversity of local data. This indicates that aggregating knowledge across nodes helps the model learn more robust object representations, especially for classes that may be unevenly distributed across the federated partitions.

The Federated model also remains close to the Centralized setting for most classes. In some cases, it even slightly outperforms the Centralized model at mAP@0.50, such as for Artillery, Soldier, Tank, and Vehicle (see Figure~\ref{fig:per_class_results}). For the stricter mAP@0.50:0.95 metric, Centralized training generally achieves the best performance, although the Federated model remains competitive. Overall, these per-class results reinforce the conclusion that FL provides a strong privacy-preserving alternative to centralized training, achieving substantial gains over isolated Single-drone learning, while maintaining competitive class-level detection performance.

\subsubsection{FL Deployment under Limited Bandwidth Conditions (600 Kbps)}

FL enables collaborative model training across distributed devices while preserving data privacy. However, deployments in wireless mesh networks and other resource-constrained environments may be affected by limited communication capacity. To evaluate the FL robustness under these conditions, we extended the previously described KIIT-MiTA experiment, using the same dataset, training scenarios, and Ultralytics YOLO11 nano model, the smallest model variant and therefore the most suitable for deployment in a drone setup.

In addition to the unconstrained centralized and federated setups shown before, we introduced a bandwidth-constrained configuration with a bidirectional cap of 600~Kbps per node, enforced through a traffic shaper. The cap was chosen as a conservative low-bandwidth stress test, motivated by prior work showing that wireless and edge FL deployments are often limited by communication capacity and resource allocation constraints~\cite{xu2020client,yao2024wireless}. To further approximate realistic drone and edge conditions, training time-related experiments were executed in a CPU-only setting, reflecting deployments where onboard computational resources are limited.

\begin{figure}[h]
\centering
  \includegraphics[scale=0.5]{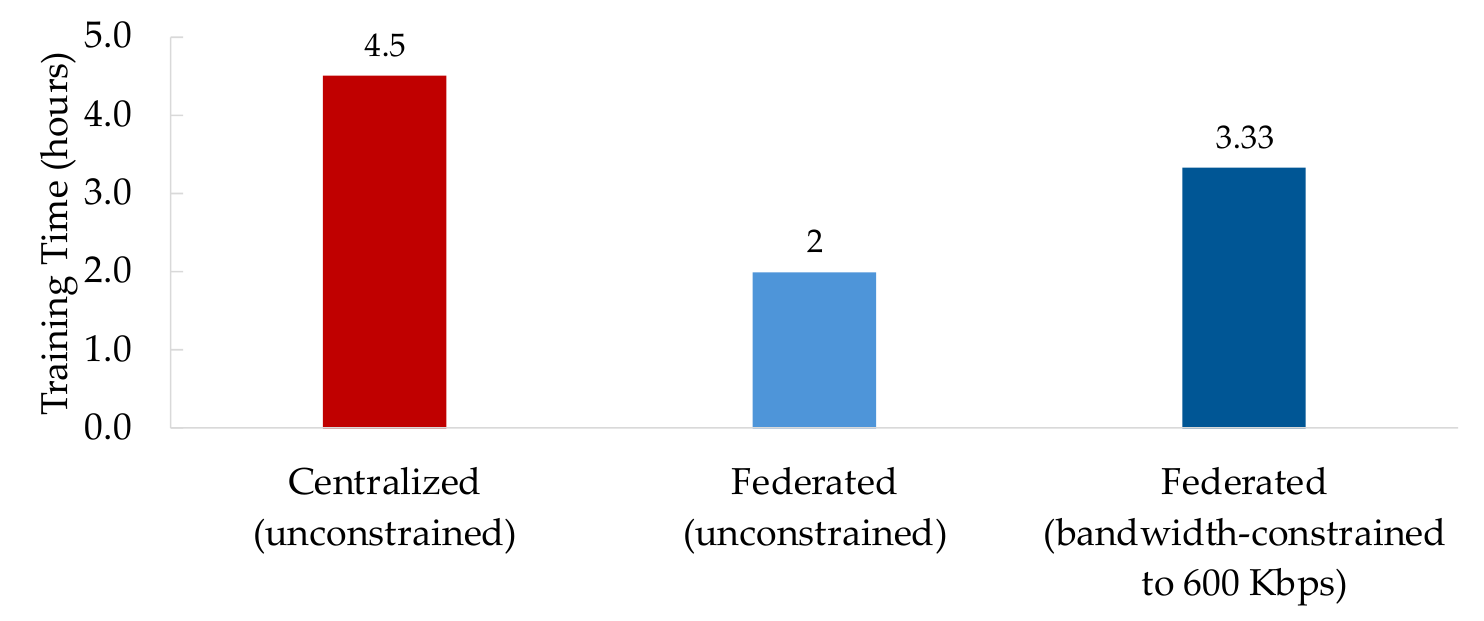}
  \caption{Training time in hours for the YOLO11 Nano model across centralized unconstrained, federated unconstrained, and federated bandwidth-constrained (600 Kbps) scenarios. 
  }\label{fig:times_training}
\end{figure}

The results show that the performance remained comparable to the full-bandwidth baseline, indicating that limited connectivity had a negligible impact on model performance. The constrained run completed federated training in 3~hours~20~minutes, compared to 2~hours under full bandwidth (see Figure~\ref{fig:times_training}). This represents a moderate increase in execution time due to the imposed bandwidth limitation and the large amount of aggregated parameters (i.e.,  around 16 MB in size) that each node needs to upload and download at each FL round. However, the execution time remained practical for real-world deployments where communication resources are limited. Importantly, both federated configurations were faster than the centralized setup, which required 4~hours~30~minutes, while also avoiding the transfer of raw data to a central server. This is particularly relevant in distributed environments, where moving large datasets can be impractical, time-consuming, and undesirable due to bandwidth, privacy, and data governance constraints.

\section{Discussion}
\label{sec:discussion}

The results show that FL can recover a large part of the benefit of multi-source training while avoiding raw-image centralization. This is important in drone deployments where visual data may be sensitive, bandwidth-limited, or distributed across different operators and locations. In this setting, FL provides a practical compromise between isolated Single-drone training, which limits generalization, and Centralized training, which requires transferring all images to a shared server.

From a deployment perspective, the main advantage of the proposed approach is that inference remains unchanged after training. Once the global model is obtained, each drone or edge node can run the detector locally with the same computational profile as a centrally trained model. Therefore, the FL strategy does not introduce inference-time overhead. However, it does introduce training-time communication costs, since model parameters must be exchanged between nodes and the server at every communication round. In real deployments, this cost should be analyzed together with network availability, model size, update frequency, and the number of participating drones.

The bandwidth-constrained experiment further illustrates this trade-off. Under a bidirectional cap of 600~Kbps per node, the Federated setting preserved a detection performance comparable to the full-bandwidth baseline, suggesting that the proposed approach can tolerate severe communication constraints without a meaningful loss in accuracy. The main effect of the bandwidth limitation was an increase in training time, from 2~hours in the unconstrained Federated setting to 3~hours~20~minutes in the constrained setting. This increase is expected, since each communication round requires nodes to upload and download model parameters. Nevertheless, the constrained Federated run remained faster than the Centralized CPU-only setup and avoided the transfer of raw images, reinforcing the suitability of FL for drone and edge deployments where connectivity is limited.

Although FL avoids exchanging raw images, it should not be interpreted as a complete privacy guarantee by itself. Model updates may still leak information about local data under certain attacks. For high-security or operationally sensitive applications, FL should therefore be combined with additional protections such as secure aggregation or differential privacy.

Our findings are also consistent with the work of Chakrabarty et al.~\cite{chakrabarty2025drones}, who showed that recent YOLO-based detectors are effective for real-time military target detection and tracking on the KIIT-MiTA dataset. While their study focuses on centralized detection and tracking pipelines, our work evaluates the same application context under Centralized, Single-drone, and Federated training scenarios. Within our experimental setup, YOLO26 nano achieved the strongest accuracy-efficiency trade-off among the evaluated models, suggesting that newer lightweight YOLO variants can further improve KIIT-MiTA object detection while preserving real-time inference capability. This comparison should be interpreted with care, since differences in model versions, training setup, and evaluation protocol may affect direct numerical comparability.

The current experiments also have limitations. First, the evaluation uses four federated nodes and a fixed non-IID partitioning strategy. Larger drone fleets may introduce stronger heterogeneity, unbalanced node participation, unreliable connectivity, and more diverse sensing conditions. Second, the experiments are conducted only on the KIIT-MiTA dataset. Although YOLO26 nano achieved the best accuracy-efficiency trade-off in this study, its advantage may depend on the dataset, object classes, image resolution, and training configuration. Additional validation on other UAV and aerial object-detection datasets is required to determine whether the same conclusions generalize beyond KIIT-MiTA.

\section{Conclusions}
\label{sec:Conclusions}

Drone-based object detection has become increasingly important for safety-critical and edge-vision applications, such as defense applications, disaster response, operational security environments, and infrastructure monitoring, where models must operate reliably across heterogeneous and dynamic environments. Achieving robust performance typically requires access to diverse aerial imagery collected across multiple drones, locations, and operating conditions. However, in real deployments, such data is often distributed, sensitive, costly to transmit, or subject to privacy and regulatory constraints, making centralized training difficult or undesirable.

Motivated by this challenge, we evaluated FL as a privacy-preserving approach for drone-based object detection on the KIIT-MiTA dataset. The experiments compared Single-drone, Centralized, and Federated training across four YOLO nano architectures under a common evaluation protocol. This setup allowed us to assess whether collaborative training can improve over isolated local learning while approaching the performance of centralized training without requiring raw image exchange.

The results show that Federated training consistently improves over the Single-drone baseline across the evaluated models and detection metrics. This confirms that collaboration among distributed nodes enables the model to benefit from a broader and more diverse training distribution than the one available to each node. At the same time, the Federated models remain close to their Centralized counterparts, indicating that much of the value of multi-source training can be preserved while keeping data local.

Among the evaluated architectures, YOLO26 nano offered the strongest overall balance between detection accuracy and computational efficiency. It achieved the best predictive performance while also requiring the lowest number of GFLOPs, making it particularly suitable for drone and edge-vision scenarios where inference speed and resource constraints are important. The per-class results further showed that the benefits of FL are not limited to the aggregate mAP values, but are reflected across all object categories.

Overall, the findings demonstrate that FL is a practical and competitive alternative to centralized training for aerial object detection when data sharing is constrained by privacy, security, bandwidth, or operational considerations. Rather than relying on transferring raw drone imagery to a central server, the proposed approach enables collaborative model improvement while preserving data locality. This makes FL a promising direction for scalable object detection in distributed drone fleets and other privacy-sensitive edge AI deployments.

% \daniel{The text added in the previous section should be revised and rewritten once the final results are available. DONE.}

\section*{Contributions and Acknowledgments} \label{app:A}

% Alex Acero

Daniel M. Jimenez-Gutierrez

Enrique Zuazua

Georgios Kellaris

Joaquin del Rio

Oleksii Sliusarenko

Xabi Uribe-Etxebarria

\vspace{5mm}
The authors are presented in alphabetical order by first name.

%\begin{appendices}

%\end{appendices}

%%%%%%%%%%%%%%%%%%%%%%%%%%%%%%%%%%%%%%%%%%%%%%%%%%%%

%%%%%%%%%%%%%%%%%%%%%%%%%%%%%%%%%%%%%%%%%%%%%%%%%%%%
%%% Bibliography
%%%%%%%%%%%%%%%%%%%%%%%%%%%%%%%%%%%%%%%%%%%%%%%%%%%%
\printbibliography %Prints bibliography

@article{ren2015faster,
  title={Faster r-cnn: Towards real-time object detection with region proposal networks},
  author={Ren, Shaoqing and He, Kaiming and Girshick, Ross and Sun, Jian},
  journal={Advances in neural information processing systems},
  volume={28},
  year={2015}
}

@inproceedings{redmon2016you,
  title={You only look once: Unified, real-time object detection},
  author={Redmon, Joseph and Divvala, Santosh and Girshick, Ross and Farhadi, Ali},
  booktitle={Proceedings of the IEEE conference on computer vision and pattern recognition},
  pages={779--788},
  year={2016}
}

@article{redmon2018yolov3,
  title={Yolov3: An incremental improvement},
  author={Redmon, Joseph and Farhadi, Ali},
  journal={arXiv preprint arXiv:1804.02767},
  year={2018}
}

@inproceedings{lin2017focal,
  title={Focal loss for dense object detection},
  author={Lin, Tsung-Yi and Goyal, Priya and Girshick, Ross and He, Kaiming and Doll{\'a}r, Piotr},
  booktitle={Proceedings of the IEEE international conference on computer vision},
  pages={2980--2988},
  year={2017}
}

@article{zhao2019object,
  title={Object detection with deep learning: A review},
  author={Zhao, Zhong-Qiu and Zheng, Peng and Xu, Shou-tao and Wu, Xindong},
  journal={IEEE transactions on neural networks and learning systems},
  volume={30},
  number={11},
  pages={3212--3232},
  year={2019},
  publisher={IEEE}
}

@misc{gdpr2016,
  title = {Regulation (EU) 2016/679 of the European Parliament and of the Council of 27 April 2016 on the protection of natural persons with regard to the processing of personal data and on the free movement of such data, and repealing Directive 95/46/EC (General Data Protection Regulation)},
  journaltitle = {Official Journal of the European Union},
  date = {2016-04-27},
  volume = {L119},
  pages = {1--88},
  pagination = {page},
  number = {L 119/1},
  url = {https://eur-lex.europa.eu/legal-content/EN/TXT/PDF/?uri=CELEX:32016R0679}
}

@misc{hipaa1996,
  title = {Health Insurance Portability and Accountability Act of 1996},
  shorttitle = {HIPAA},
  number = {Pub. L. No. 104-191},
  pages = {110 Stat. 1936},
  date = {1996-08-21},
  pagination = {section}
}

@misc{CCPA2018,
  title = {California Consumer Privacy Act},
  journaltitle = {California Civil Code},
  date = {2018},
  number = {1798.100-1798.199},
  pagination = {section},
}

@misc{lopd2018,
  title = {Ley Orgánica 3/2018, de 5 de diciembre, de Protección de Datos Personales y garantía de los derechos digitales},
  journaltitle = {Boletín Oficial del Estado},
  date = {2018-12-06},
  number = {294},
  pages = {119788-119857},
  url = {https://www.boe.es/eli/es/lo/2018/12/05/3},
}

@inproceedings{mcmahan2017communication,
  title={Communication-efficient learning of deep networks from decentralized data},
  author={McMahan, Brendan and Moore, Eider and Ramage, Daniel and Hampson, Seth and y Arcas, Blaise Aguera},
  booktitle={Artificial intelligence and statistics},
  pages={1273--1282},
  year={2017},
  organization={PMLR}
}

@article{yang2019federated,
  title={Federated machine learning: Concept and applications},
  author={Yang, Qiang and Liu, Yang and Chen, Tianjian and Tong, Yongxin},
  journal={ACM Transactions on Intelligent Systems and Technology (TIST)},
  volume={10},
  number={2},
  pages={1--19},
  year={2019},
  publisher={ACM New York, NY, USA}
}

@inproceedings{girshick2014rich,
  title={Rich feature hierarchies for accurate object detection and semantic segmentation},
  author={Girshick, Ross and Donahue, Jeff and Darrell, Trevor and Malik, Jitendra},
  booktitle={Proceedings of the IEEE conference on computer vision and pattern recognition},
  pages={580--587},
  year={2014}
}

@inproceedings{girshick2015fast,
  title={Fast r-cnn},
  author={Girshick, Ross},
  booktitle={Proceedings of the IEEE international conference on computer vision},
  pages={1440--1448},
  year={2015}
}

@inproceedings{tian2019fcos,
  title={Fcos: Fully convolutional one-stage object detection},
  author={Tian, Zhi and Shen, Chunhua and Chen, Hao and He, Tong},
  booktitle={Proceedings of the IEEE/CVF international conference on computer vision},
  pages={9627--9636},
  year={2019}
}

@inproceedings{carion2020end,
  title={End-to-end object detection with transformers},
  author={Carion, Nicolas and Massa, Francisco and Synnaeve, Gabriel and Usunier, Nicolas and Kirillov, Alexander and Zagoruyko, Sergey},
  booktitle={European conference on computer vision},
  pages={213--229},
  year={2020},
  organization={Springer}
}

@inproceedings{lin2014microsoft,
  title={Microsoft coco: Common objects in context},
  author={Lin, Tsung-Yi and Maire, Michael and Belongie, Serge and Hays, James and Perona, Pietro and Ramanan, Deva and Doll{\'a}r, Piotr and Zitnick, C Lawrence},
  booktitle={Computer vision--ECCV 2014: 13th European conference, zurich, Switzerland, September 6-12, 2014, proceedings, part v 13},
  pages={740--755},
  year={2014},
  organization={Springer}
}

@article{everingham2010pascal,
  title={The pascal visual object classes (voc) challenge},
  author={Everingham, Mark and Van Gool, Luc and Williams, Christopher KI and Winn, John and Zisserman, Andrew},
  journal={International journal of computer vision},
  volume={88},
  pages={303--338},
  year={2010},
  publisher={Springer}
}

@article{caldas2018leaf,
  title={Leaf: A benchmark for federated settings},
  author={Caldas, Sebastian and Duddu, Sai Meher Karthik and Wu, Peter and Li, Tian and Kone{\v{c}}n{\`y}, Jakub and McMahan, H Brendan and Smith, Virginia and Talwalkar, Ameet},
  journal={arXiv preprint arXiv:1812.01097},
  year={2018}
}

@article{sheller2020federated,
  title={Federated learning in medicine: facilitating multi-institutional collaborations without sharing patient data},
  author={Sheller, Micah J and Edwards, Brandon and Reina, G Anthony and Martin, Jason and Pati, Sarthak and Kotrotsou, Aikaterini and Milchenko, Mikhail and Xu, Weilin and Marcus, Daniel and Colen, Rivka R and others},
  journal={Scientific reports},
  volume={10},
  number={1},
  pages={12598},
  year={2020},
  publisher={Nature Publishing Group UK London}
}

@article{li2020federated,
  title={Federated learning: Challenges, methods, and future directions},
  author={Li, Tian and Sahu, Anit Kumar and Talwalkar, Ameet and Smith, Virginia},
  journal={IEEE signal processing magazine},
  volume={37},
  number={3},
  pages={50--60},
  year={2020},
  publisher={IEEE}
}

@inproceedings{liu2016ssd,
  title={Ssd: Single shot multibox detector},
  author={Liu, Wei and Anguelov, Dragomir and Erhan, Dumitru and Szegedy, Christian and Reed, Scott and Fu, Cheng-Yang and Berg, Alexander C},
  booktitle={Computer Vision--ECCV 2016: 14th European Conference, Amsterdam, The Netherlands, October 11--14, 2016, Proceedings, Part I 14},
  pages={21--37},
  year={2016},
  organization={Springer}
}

@article{lu2024federated,
  title={Federated learning with non-iid data: A survey},
  author={Lu, Zili and Pan, Heng and Dai, Yueyue and Si, Xueming and Zhang, Yan},
  journal={IEEE Internet of Things Journal},
  year={2024},
  publisher={IEEE}
}

@article{zhu2021federated,
  title={Federated learning on non-IID data: A survey},
  author={Zhu, Hangyu and Xu, Jinjin and Liu, Shiqing and Jin, Yaochu},
  journal={Neurocomputing},
  volume={465},
  pages={371--390},
  year={2021},
  publisher={Elsevier}
}

@inproceedings{liu2020fedvision,
  title={Fedvision: An online visual object detection platform powered by federated learning},
  author={Liu, Yang and Huang, Anbu and Luo, Yun and Huang, He and Liu, Youzhi and Chen, Yuanyuan and Feng, Lican and Chen, Tianjian and Yu, Han and Yang, Qiang},
  booktitle={Proceedings of the AAAI conference on artificial intelligence},
  volume={34},
  number={08},
  pages={13172--13179},
  year={2020}
}

@article{zhu2018vision,
  title={Vision meets drones: A challenge},
  author={Zhu, Pengfei and Wen, Longyin and Bian, Xiao and Ling, Haibin and Hu, Qinghua},
  journal={arXiv preprint arXiv:1804.07437},
  year={2018}
}

@misc{chakrabarty2025kiitmita,
  author       = {Sudip Chakrabarty},
  title        = {KIIT-MiTA},
  year         = {2025},
  howpublished = {\url{https://www.kaggle.com/datasets/sudipchakrabarty/kiit-mita}},
  note         = {Kaggle dataset, accessed 2026-04-22}
}

@article{leng2024recent,
  title={Recent advances for aerial object detection: A survey},
  author={Leng, Jiaxu and Ye, Yongming and Mo, Mengjingcheng and Gao, Chenqiang and Gan, Ji and Xiao, Bin and Gao, Xinbo},
  journal={ACM Computing Surveys},
  volume={56},
  number={12},
  pages={1--36},
  year={2024},
  publisher={ACM New York, NY}
}

@inproceedings{chakrabarty2025drones,
  title={Drones in defense: Real-time vision-based military target surveillance and tracking},
  author={Chakrabarty, Sudip and Chatterjee, Rajdeep and Chakraborty, Sorup and Shuvo, Sourov Roy and Chowdhury, Rajesh},
  booktitle={2025 3rd International Conference on Intelligent Systems, Advanced Computing and Communication (ISACC)},
  pages={508--513},
  year={2025},
  organization={IEEE}
}

@article{lu2025improving,
  title={Improving Federated Learning UAV Urban Object Detection System via Data Heterogeneity Mitigation},
  author={Lu, You-Ru and Sun, Dengfeng},
  journal={Journal of Aerospace Information Systems},
  volume={22},
  number={11},
  pages={930--937},
  year={2025},
  publisher={American Institute of Aeronautics and Astronautics}
}

@article{jimenez2024fedartml,
  title={FedArtML: A Tool to Facilitate the Generation of Non-IID Datasets in a Controlled Way to Support Federated Learning Research},
  author={Jimenez-Gutierrez, Daniel M and Anagnostopoulos, Aris and Chatzigiannakis, Ioannis and Vitaletti, Andrea},
  journal={IEEE Access},
  year={2024},
  publisher={IEEE}
}

@article{lu2024development,
  title={Development of real-time unmanned aerial vehicle urban object detection system with federated learning},
  author={Lu, You-Ru and Sun, Dengfeng},
  journal={Journal of Aerospace Information Systems},
  volume={21},
  number={7},
  pages={547--553},
  year={2024},
  publisher={American Institute of Aeronautics and Astronautics}
}

@inproceedings{wang2023generalized,
  title={Generalized uav object detection via frequency domain disentanglement},
  author={Wang, Kunyu and Fu, Xueyang and Huang, Yukun and Cao, Chengzhi and Shi, Gege and Zha, Zheng-Jun},
  booktitle={Proceedings of the IEEE/CVF conference on computer vision and pattern recognition},
  pages={1064--1073},
  year={2023}
}

@software{yolov5_ultralytics,
  title = {Ultralytics YOLOv5},
  author = {Jocher, Glenn},
  year = {2020},
  version = {7.0},
  license = {AGPL-3.0},
  url = {https://github.com/ultralytics/yolov5},
  doi = {10.5281/zenodo.3908559},
  orcid = {0000-0001-5950-6979}
}

@software{yolov8_ultralytics,
  author = {Jocher, Glenn and Chaurasia, Ayush and Qiu, Jing},
  title = {Ultralytics YOLOv8},
  version = {8.0.0},
  year = {2023},
  url = {https://github.com/ultralytics/ultralytics},
  orcid = {0000-0001-5950-6979, 0000-0002-7603-6750, 0000-0003-3783-7069},
  license = {AGPL-3.0}
}

@software{yolo11_ultralytics,
  author = {Jocher, Glenn and Qiu, Jing},
  title = {Ultralytics YOLO11},
  version = {11.0.0},
  year = {2024},
  url = {https://github.com/ultralytics/ultralytics},
  orcid = {0000-0001-5950-6979, 0000-0003-3783-7069},
  license = {AGPL-3.0}
}

@software{yolo26_ultralytics,
  author = {Jocher, Glenn and Qiu, Jing},
  title = {Ultralytics YOLO26},
  version = {26.0.0},
  year = {2026},
  url = {https://github.com/ultralytics/ultralytics},
  orcid = {0000-0001-5950-6979, 0000-0003-3783-7069},
  license = {AGPL-3.0}
}

@software{ultralytics_framework,
  author = {Jocher, Glenn and Qiu, Jing and Chaurasia, Ayush},
  title = {Ultralytics YOLO},
  year = {2023},
  url = {https://github.com/ultralytics/ultralytics},
  license = {AGPL-3.0}
}

@article{xu2020client,
  title={Client selection and bandwidth allocation in wireless federated learning networks: A long-term perspective},
  author={Xu, Jie and Wang, Heqiang},
  journal={IEEE Transactions on Wireless Communications},
  volume={20},
  number={2},
  pages={1188--1200},
  year={2020},
  publisher={IEEE}
}

@article{yao2024wireless,
  title={Wireless federated learning over resource-constrained networks: Digital versus analog transmissions},
  author={Yao, Jiacheng and Xu, Wei and Yang, Zhaohui and You, Xiaohu and Bennis, Mehdi and Poor, H Vincent},
  journal={IEEE Transactions on Wireless Communications},
  volume={23},
  number={10},
  pages={14020--14036},
  year={2024},
  publisher={IEEE}
}

\end{document}